\documentclass[10pt,a4paper,oneside]{article}
\usepackage[T1]{fontenc}
\usepackage[utf8]{inputenc}
\usepackage{tikz}
\usetikzlibrary{shapes,arrows}
\tikzstyle{block} = [rectangle, draw, fill=blue!20, 
    text width=5em, text centered, rounded corners, minimum height=4em]
\usepackage{geometry}  
\usepackage[american]{babel}
\usepackage[babel]{csquotes}
\usepackage{amssymb}

\usepackage[%
            style=numeric-comp,%
            sorting=nyt,%
            abbreviate=true,%
            useprefix=true,%
            giveninits=true,%
            maxcitenames=3,%
            maxbibnames=20,%
            uniquename=init,%
            sortcites=true,%
            doi=true,%
            isbn=false,%
            url=false,%
            eprint=false,%
            backend=biber%
                           ]{biblatex} %
  \renewbibmacro{in:}{}
  \DefineBibliographyStrings{english}{bibliography = {References}}
  \preto\fullcite{\AtNextCite{\defcounter{maxnames}{99}}}
  \AtEveryBibitem{\clearfield{month}} %
  \AtEveryCitekey{\clearfield{month}}

\usepackage{enumitem}
\usepackage[version=3]{mhchem} %
\usepackage[font=scriptsize]{caption}
\usepackage{graphicx}
\usepackage{booktabs}  %
\usepackage{tabularx}  %
  \newcolumntype{R}{>{\raggedleft\arraybackslash}X}
  \newcolumntype{L}{>{\raggedright\arraybackslash}X}
  \newcolumntype{C}{>{\centering\arraybackslash}X}
      
\usepackage{hyperref}
\hypersetup{
final,                     %
pdftoolbar=true,           %
pdfmenubar=true,           %
pdftitle={Estimating permeability of 3D~micro-CT images by physics-informed CNNs based on DNS},     %
pdfauthor={G\"arttner et al.},  %
pdfnewwindow=true,         %
colorlinks=true,           %
linkcolor=[rgb]{0,0.3,0.3},            %
citecolor=[rgb]{0,0.3,0.3},            %
filecolor=[rgb]{0,0.3,0.3},            %
urlcolor=[rgb]{0,0.3,0.3}              %
}

\usepackage{amsthm}
\usepackage{cleveref} %
  \crefformat{equation}{(#2#1#3)}
  \Crefformat{equation}{Equation~#2#1#3}
  \crefformat{figure}{Figure~#2#1#3}         \crefname{figure}{Figure}{Figures}
  \crefformat{section}{Section~#2#1#3}      \crefname{section}{Section}{Sections}
  \Crefformat{section}{Section~#2#1#3}      \Crefname{section}{Section}{Sections}
  \crefformat{table}{Table~#2#1#3}          \crefname{table}{Table}{Tables}
  \Crefformat{table}{Table~#2#1#3}          \Crefname{table}{Table}{Tables}

\usepackage{mathtools}
\usepackage[OMLmathsfit]{isomath} %

\usepackage{siunitx}                 %
\newcommand*{\muCT}{\textmu CT}
\renewcommand*{\vec}[1]{{\boldsymbol{#1}}}                       %
\DeclareMathAlphabet{\mathbfsf}{\encodingdefault}{\sfdefault}{bx}{n}
  \newcommand*{\vecc}[1]{\mathbfsf{#1}}                          %
\newcommand*{\normal}{\vec{n}}                                   %
\newcommand*{\laplace}{\upDelta}                                 %
\newcommand*{\grad}{\vec{\nabla}}                                %
\renewcommand*{\div}{\vec{\nabla}\cdot}                          %
\newcommand*{\tp}{\mathrm{T}}                                    %
\newcommand*{\dd}{\mathrm{d}}  %
\DeclareMathOperator{\card}{card} %

\usepackage{dsfont}

\newcommand{\IP}{\ensuremath\mathds{P}}
\newcommand{\IQ}{\ensuremath\mathds{Q}}
\newcommand{\IR}{\ensuremath\mathds{R}}
\renewcommand{\Re}{\mathrm{Re}} %
\newcommand{\Da}{\mathrm{Da}} %

\newcommand*{\numU}{n}
\newcommand*{\numP}{m}
\newcommand*{\xU}{\vec{x}^\vec{u}} 
\newcommand*{\xP}{\vec{x}^p}
\newcommand*{\bU}{\vec{b}^\vec{u}}

\usepackage[symbol]{footmisc} %

\newcommand*{\orcid}[1]{\href{https://orcid.org/#1}{\includegraphics[width=1em]{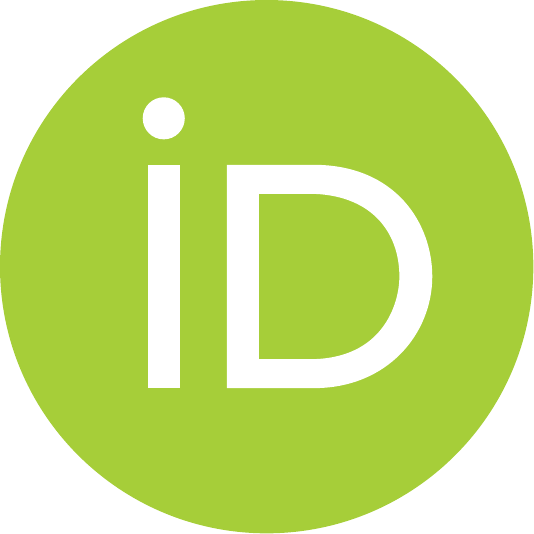}}}
\title{Estimating permeability of 3D~micro-CT images by physics-informed CNNs based on DNS}
\author{Stephan G\"arttner\textsuperscript{1,*}\orcid{0000-0002-3488-0929},
Faruk O.~Alpak\textsuperscript{2}\orcid{0000-0002-0738-6298},
Andreas Meier\textsuperscript{1}\orcid{0000-0002-3135-9310}, 
\\
Nadja Ray\textsuperscript{1}\orcid{0000-0002-9596-953X},
Florian Frank\textsuperscript{1,*}\orcid{0000-0002-9756-1351}
\\
\normalsize\begin{tabular}{rl}
{}\\
\textsuperscript{1} & Friedrich-Alexander-Universität Erlangen-Nürnberg, \\
{} & Department~Mathematik, Cauerstraße~11, 91058~Erlangen, Germany
\\
\textsuperscript{2} & Shell~Technology~Center, 3333~Highway~6~South, Houston, TX~77082, USA
\end{tabular}
}
\date{\today}

\begin{document}

\maketitle
\thispagestyle{empty}

\footnotetext{\textsuperscript{*} Corresponding authors. \emph{E-mail address:} \url{gaerttner@math.fau.de}, \url{frank@math.fau.de}.}



\begin{abstract}
In recent years, convolutional neural networks (CNNs) have experienced an~increasing interest for their ability to perform fast approximation of effective hydrodynamic parameters in porous media research and applications. This paper presents a~novel methodology for permeability prediction from micro-CT scans of geological rock samples. 
The training data set for CNNs dedicated to permeability prediction consists of permeability labels that are typically generated by classical lattice Boltzmann methods (LBM) that simulate the flow through the pore space of the segmented image data. We instead perform direct numerical simulation (DNS) by solving the stationary Stokes equation in an~efficient and distributed-parallel manner. As such, we circumvent the convergence issues of LBM that frequently are observed on complex pore geometries, and therefore, improve on the generality and accuracy of our training data set.
\par
Using the DNS-computed permeabilities, a~physics-informed CNN \mbox{(PhyCNN)} is trained by additionally providing a~tailored characteristic quantity of the pore space. More precisely, by exploiting the connection to flow problems on a~graph representation of the pore space, additional information about confined structures is provided to the network in terms of the maximum flow value, which is the key innovative component of our workflow.  The robustness of this approach is reflected by very high  prediction accuracy, which is observed for a~variety of sandstone samples from archetypal rock formations. 
\end{abstract}

\section{Introduction}
Artificial neural networks can accelerate or replace classical methods for estimating various hydrological and petrophysical properties of artificial and natural rock~\cite{WANGReview}. In~\cite{ArayaPoloEtAl2018}, deep neural networks were successfully used to approximate tomography operators to reconstruct wave velocity models from seismic data. Moreover, convolutional neural networks (CNNs)  were used in~\cite{CnnHomo} to determine effective electrical parameters in a homogenized model for electric conductivity. Likewise, CNNs have proven their ability to provide fast predictions of scalar permeability values directly from images of the pore space of geological specimens~\cite{CNNPerm,RapidEstimate,CNNPrifling}. Finally, CNNs were successfully used to replace standard solution schemes to predict effective permeabilities in 2D multiscale flow simulations~\cite{Gaerttner2020b}. This paper aims at contributing to the deployment of machine learning techniques in effective permeability prediction by presenting a~finite-element-based forward simulation approach as well as introducing a~novelly considered characteristic quantity for physics-informed neural network~\mbox{(PhyCNN)} models. We use the term PhyCNN throughout this paper to underline the incorporation of an additional physically motivated input parameter to the network. This approach is to be carefully distinguished from implementing physical laws into the loss-function as performed in~\cite{ChengPhyLoss2021} to estimate flow velocity fields.
\par
Specimens of natural rock are typically obtained by microcomputed~tomography~(\muCT) scanning~\cite{WildenschildSheppard2013,BultreysBoeverCnudde2016} or indirectly from 2D~colored images~\cite{ArayaPoloEtAl2019}. As an~important characteristic quantity of porous media, permeability measures the ability of a~fluid to travel through a~considered pore space. However, in general, the precision and generality of the estimates performed by neural networks heavily depend on the quality of the underlying training data set, i.\,e., the accuracy of the forward simulations in our case, as shown in~\cite{ZhouDataSetQuality}. In terms of permeability prediction, the labeling process of geological specimens is connected to the computation of 3D~stationary flow fields of a~single-phase fluid within the pore space. For permeability prediction under regular geometries, a~broader set of methods is available as described and compared in~\cite{RybakPermeabilityEstimation2021}.
\par
In the literature, various methods are available to compute the stationary flow on complex geometries as constituted by the pore space of porous media, a~thorough comparison of which is found in~\cite{YANG2016176}. Most commonly, lattice Boltzmann methods (LBM) are used to solve the flow problem on a~discrete modeling basis, cf.~\cite{LBMStokes}. In this approach, a~transient interacting many-particle system is driven to equilibrium state, heavily exploiting the inherent parallelism of the underlying mathematical structure. For a~detailed description of LBM fundamentals and numerics, we refer to~\cite{BookKruegerEtAl2016}. Yet, since a~global equilibrium has to be reached, complex geometries containing thin channels may cause LBM to converge slowly or even diverge~\cite{SantosEtAl2020,RapidEstimate}, especially in the case of very constricted and therefore typically under-resolved pore morphologies. This behavior is frequently observed with common pore-scale flow LBM implementations that rely on the conventional single-relaxation-time Bhatnagar--Gross--Krook (BGK) scheme~\cite{BhatnagarEtAl1954}. However, the novel multiple-relaxation-time (MRT) scheme addresses this problem to a~great extent (e.\,g.,~\cite{AlpakLBM2018}) at the cost of a~moderate additional computational overhead. Having stated that, residual discretization errors stemming from the explicit nature of time integration in LBM schemes still remain in the numerical solution. Especially for under-resolved pore morphologies, these errors may be amplified causing divergence of the numerical flow solution. We further note that classical bounce-back rules used for implementing boundary conditions tend to develop oscillations that might locally dominate fluid behavior in scenarios with thin channels, cf.~\cite{YangLBMboundary}. Severe limitations arising from simplistic implementations of LBM restrict the training data sets to ones derived from sphere-packs instead of real CT~data in many publications, see for instance~\cite{SantosEtAl2020}. On the other hand, industrial-grade state-of-the-art LBM solvers are typically proprietary without public access to the code base for academic research purposes (e.\,g.,~\cite{Toelke2008,AlpakLBM2018}). Moreover, a~comprehensive permeability computation benchmarking study~\cite{Saxena2017} demonstrated that not all of the simulation methods deliver a~good compromise of accuracy against computational performance indicating that there exists a~clear need for accurate and computationally efficient alternative methods for permeability computation. 
\par
In practice, LBM simulations are often aborted after a~maximal number of iterations in case one or more convergence criteria (typically linked to the relative changes in the velocity and/or computed permeability) cannot be met~\cite{RapidEstimate}. As such, training sets for neural networks can be artificially filtered by the numerical characteristics of the forward simulation, potentially resulting in biased data. To improve the generality and quality of our \mbox{PhyCNN} training sets, we base our machine learning data set on the stationary Stokes equation by performing direct numerical simulations (DNS) on the pore geometry~\cite{DNS}. More precisely, our forward simulation is based on a~distributed-parallel Stokes solver utilizing the finite element library~MFEM~\cite{mfem}. As studied in~\cite{RybakPermeabilityEstimation2021} for simple cylindrical obstacles, such DNS approaches (in this case FEM) deliver permeability values comparable to the ones from LBM simulation. However, our implementation successfully alleviates the drawback of an~impractically large number of iterations to obtain the desired accuracy on complex geometries. As such, our approach allows overcoming prior restrictions in setting up representative training sets including also confined and complex structures. Moreover, no artificial data augmentation schemes such as pore space dilation are needed in our approach to increase the number of training data samples or enhance the porosity range covered. 
\par
Likewise, novel strategies are employed to our neural networks. 
More precisely, we exploit the concept of \mbox{PhyCNNs}, where the CNN is provided with additional (physics-related) input quantities to improve the reliability and the accuracy of its predictions~\cite{CNNPerm, CNNTEMBELY}. As shown in~\cite{sun2019epci}, carefully chosen specific quantities derived from the pore space such as connectivity indices can deliver reasonable approximation quality for permeability estimation. As illustrated in~\cite{ArayaPoloEtAl2018}, also training a~network solely on previously extracted features from the raw data may lead to satisfactory prediction quality.
\par
The outstanding performance of our methodology is achieved by considering the \emph{maximum flow} value, a~graph-network derived quantity being highly correlated to the target permeability value and simple to compute. As such, we solve maximum flow problems on a~graph representation of the pore space based on~\cite{MaxFlow}. Thus, we approximate the Stokes flow through the pore space by an abstract flow through a~graph. By using the scalar quantity of maximum flow as a~second input to our neural network, we additionally provide our CNN with information that reflects possible thin, channel-like structures. As discussed in~\cite{GraphAlgoFlow}, graph representations have been shown highly capable of characterizing the pore-space connectivity in fractured rock and allow for a~convenient way to deduce topological quantities of interest. We demonstrate that involving the maximum flow value in our newly designed \mbox{PhyCNN} in combination with the DNS-based forward simulation approach, our methodology delivers superior prediction accuracy and robustness compared to what is found in the literature. 
\par
The paper is organized as follows: In~\cref{SEC:Datapreparation}, we describe the sampling and preprocessing procedure of sandstone specimens including the forward simulation. Subsequently, \cref{SEC:MachineLearningModel} is dedicated to the network architecture used in our study. Finally, we validate the training performance of our PhyCNN on different types of sandstone in~\cref{SEC:CNNTrainingAndWorkflowPerformance}.

\section{Methodology and data preparation}
\label{SEC:Datapreparation}
\begin{figure}
    \centering
  \begin{tikzpicture}
    {\footnotesize
    \node (A) at (0,0) [align=center,rounded corners,rectangle split, rectangle split parts=2, minimum width = 3.2cm, minimum height = 5cm, fill=blue!8,text width = 3.2cm,draw] {
    \textbf{Data acquisition}
    \nodepart[align=left]{two}
    \begin{itemize}[leftmargin=*,topsep=0pt]
    \item Extract $100^3$-voxel subsamples.
    \item Remove dis\-con\-nect\-ed pore space.
    \item Eliminate impermeable subsamples.
    \end{itemize}    };
    
    \node (B) at (3.8,0) [align=center,rounded corners,rectangle split, rectangle split parts=2,minimum width = 3.2cm, minimum height = 5cm,fill=green!8,text width = 3.2cm,draw] {
    \textbf{Data labeling}
    \nodepart[align=left]{two}
    Perform forward simulation to determine permeability labels~$k_\mathrm{cmp}$.
    };
    
    \node (C) at (7.6,0) [align=center,rounded corners,rectangle split, rectangle split parts=2, minimum width = 3.2cm, minimum height = 5cm,fill=yellow! 10,text width = 3.2cm,draw] {
    \textbf{Network training}
    \nodepart[align=left]{two}
    \begin{itemize}[leftmargin=*,topsep=0pt]
    \item Compute physics input.
    \item Setup training and validation data set.
    \item Optimize network parameters via SGD.
    \end{itemize}    };
    
    \node (D) at (11.4,0) [align=center,rounded corners,rectangle split, rectangle split parts=2, minimum width = 3.2cm, minimum height = 5cm,fill=orange! 10,text width = 3.2cm,draw] {
    \textbf{Network validation} against
    \nodepart[align=left]{two}
    \begin{itemize}[leftmargin=*,topsep=0pt]
    \item unseen data samples of same sandstone type,
    \item data samples of different sandstone type,
    \item artificially distorted data.
    \end{itemize}    };
    
    \draw[->, black, ultra thick] (A) to (B);
    \draw[->, black, ultra thick] (B) to (C);
    \draw[->, black, ultra thick] (C) to (D);}
   
 \end{tikzpicture}
    \caption{Overall workflow in flow chart representation. }
    \label{fig:FlowChart}
\end{figure}
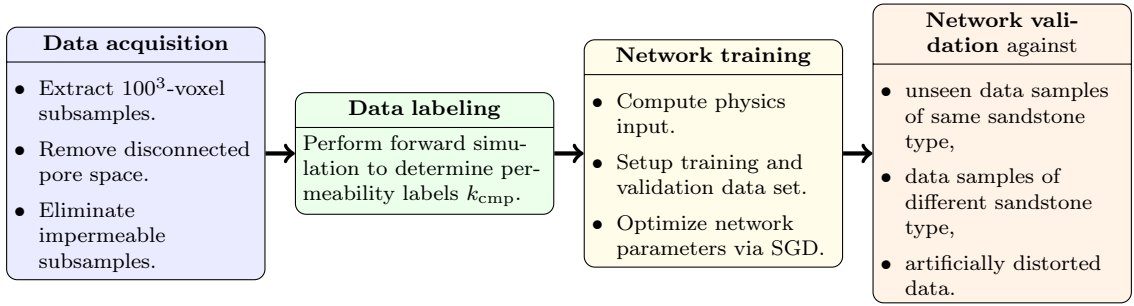

In this section, we describe the workflow and methodology by which we acquire the data set necessary to train and validate a~CNN using a~supervised learning approach. The preprocessing includes the selection and preparation of a~set of~pore-space geometries in form of voxel sets and the labeling with their computed permeability value $k_\mathrm{cmp}$. Our complete workflow is presented in~\cref{fig:FlowChart}. We note that all steps except for the data labeling procedure (green box) are implemented in Matlab~2021a~\cite{MATLAB:2021a}.

\subsection{Sampling and preprocessing} 
\label{SEC:DataSampling}
\begin{figure}[ht!]
  \centering
  \includegraphics[width=\linewidth]{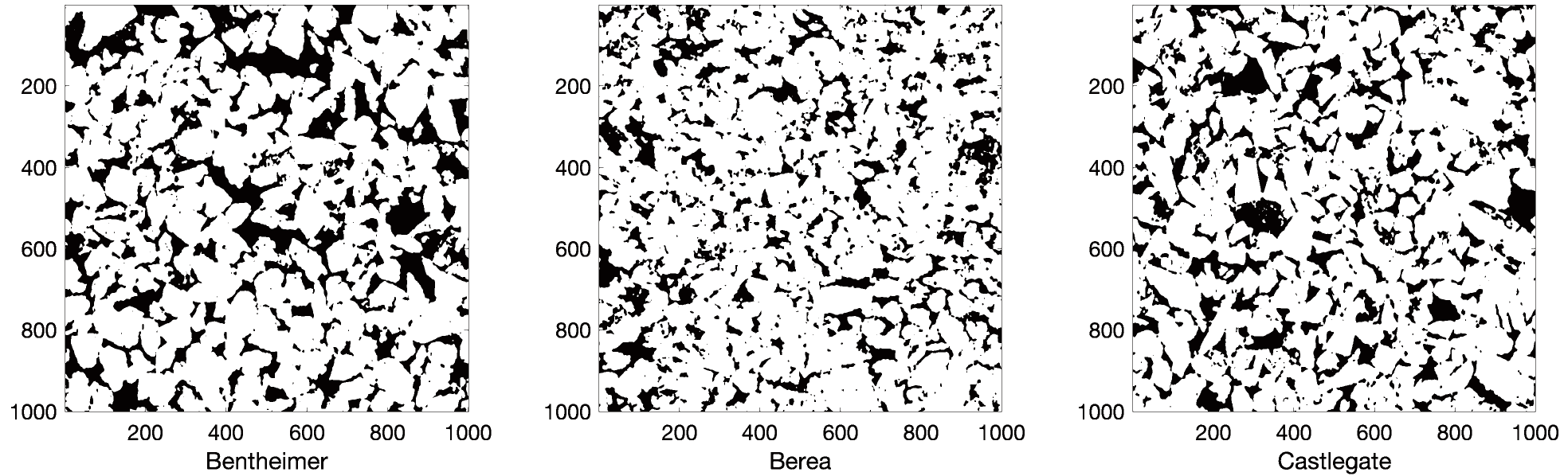}
  \\[\baselineskip]
  ~~~\includegraphics[width=0.95\linewidth]{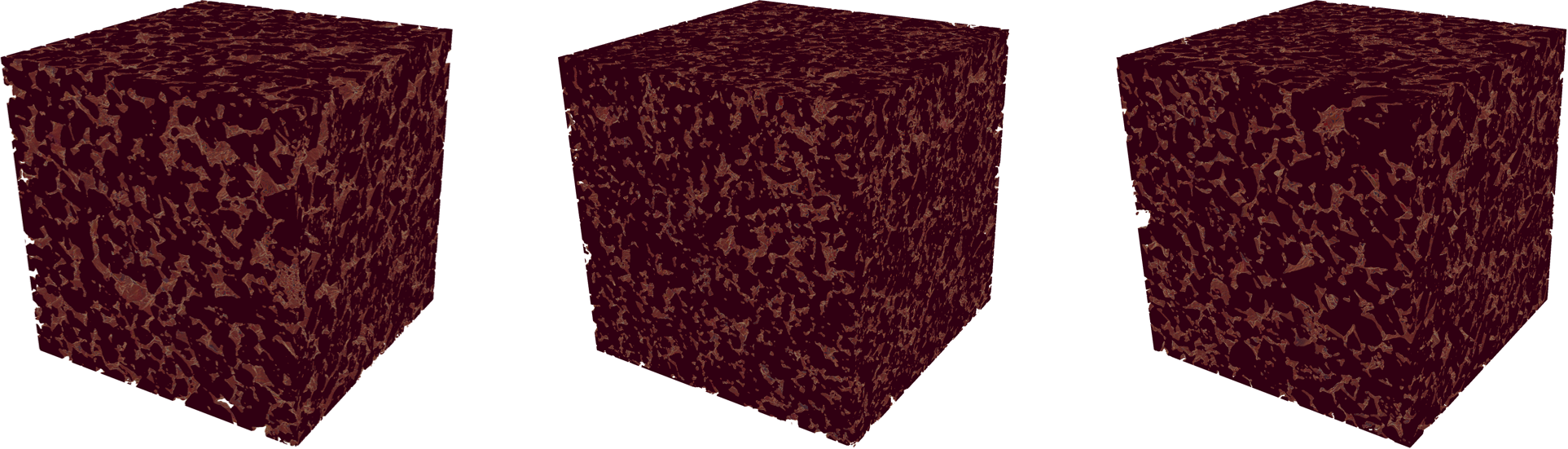}
  \caption{1000$^2$ 2D~slices and 1000$^3$ 3D~pore-space \muCT~image of all samples considered in this study, illustrating characteristic pore features for Bentheimer, Berea, and Castlegate sandstone.}
  \label{fig:PoreSpace}
\end{figure}

The training procedure of our \mbox{PhyCNN} is based on a~segmented \mbox{X-ray} \muCT~scan of a~Bentheimer sandstone sample, see~\cref{fig:PoreSpace}, with experimentally measured porosity~$\phi_\mathrm{exp}=22.64\%$ and permeability~$k_\mathrm{exp}=\SI{386}{mD}$, provided by~\cite{BentheimerDataSet,BentheimerPublication}. Further characteristic quantities for this sandstone as well as two other sandstone types used below for validation purposes are listed in~\cref{TAB:SampleCharacteristics}. Bentheimer sandstone is known to exhibit a~broad range in pore volume distribution and high pore connectivity in comparison to other types of sandstone~\cite{PoreGeoLinxian}. As such, this sample is expected to contain a~representative collection of geometrical and topological properties of the pore space in natural rocks. 
The data set used in this paper is derived from a~$\num{1000}{\times}\num{1000}{\times}\num{1000}$~binary voxel image, in which each voxel either belongs to the pore space (\enquote{fluid voxels}) or the rock matrix~(\enquote{solid voxels}). The voxel edge length is~$\SI{2.25}{\micro\meter}$ yielding an overall cube side length of $\SI{2.25}{\milli\meter}$.
\par

\begin{table}[ht!]
  \centering
  \begin{tabular}{ l c c c c c}
   \toprule
    type & $\phi_\mathrm{exp}$ & $\phi_\mathrm{cmp}$ & $k_\mathrm{exp}$ [mD] & MCD [$\si{\micro\meter}$] & $A_\mathrm{cmp}$ [$\si{\milli\meter}$]\\
    \midrule
    Bentheimer  & 22.64\% & 26.72\% & 386  & 30.0  & 355  \\
    Berea       & 18.96\% & 21.67\% & 121  & 22.3  & 284  \\
    Castlegate  & 26.54\% & 24.67\% & 269  & 24.7  & 335  \\
    \bottomrule
  \end{tabular}
  \caption{Characteristics quantities of sandstone samples used in this paper: experimentally determined porosity, $\phi_\mathrm{exp}$ ($\pm 0.5\%$), porosity computed from the \muCT~scan, $\phi_\mathrm{cmp}$, experimentally determined permeability, $k_\mathrm{exp}$ ($\pm 10\%$), mean capillary diameter, MCD, and interior surface area, $A_\mathrm{cmp}$, computed via the Matlab function \texttt{isosurface}. First four quantities are provided in~\cite{BentheimerPublication}. }
  \label{TAB:SampleCharacteristics}  
\end{table}

In the first step, we extract subsamples of $100{\times}100{\times}100$ voxels from the Bentheimer sample. Henceforth, we use the term \enquote*{subsample} to refer to segmented \muCT-scan pieces of this specific size. For subsample extraction, we make use of the sliding frame technique. This approach is commonly used to further exploit a~given data set beyond the partition into disjoint subsets~\cite{Sudakov}. More precisely, we sweep a~$100{\times}100{\times}100$ voxel frame along the coordinate axes of the original $1000$-voxel cube and displace it by steps of $50$~voxels (half a~subsample size) resulting in a~total of $19^3 = \num{6859}$ subsamples. Although data are sampled redundantly, the set of obtained subsamples can be regarded independently when training neural networks~\cite{RapidEstimate}. Furthermore, by rotating the original $100{\times}100{\times}100$ voxel subsamples by $90^\circ$ around the $y$ and $z$~axis, the number of extractable data is increased further by a~factor of three. As such, artificial data augmentation techniques like erosion and dilation of the pore space from the \muCT~image are not required here for obtaining a~sufficient and representative amount of training data. 
Even though we can produce $3{\cdot}19^3{=}\num{20577}$ subsamples by the method above, we select only the first \num{10000} ones, since this number is sufficient to train our PhyCNN. In particular, this number of available data samples exceeds that of similar studies using natural rock sample for training, cf.~\cite{RapidEstimate, SantosEtAl2020}. However, using artificially generated pore-space geometries, even larger data sets including \num{90000} data point are available in the literature, cf.~\cite{CNNPrifling}. More precisely, these data are obtained from stochastic computer models, evading the necessity of expensive imaging and segmentation of real rock.
\par
Second, fluid voxels belonging to disconnected pore space with respect to the $x$~direction possibly occurring within the subsamples are turned into solid voxels. As disconnected pores do not contributed to Stokes flow being driven from the inhomogeneous boundary conditions~\cref{EQ:StokesProblem:NeumannBC} placed on opposing sides of the subsample, this procedure maintains permeability properties while facilitating their calculation. To this end, a~simple graph walking algorithm is exploited to identify connected subdomains of the pore space. Starting from a~random fluid node, neighboring nodes within the fluid domain are successively added until the scheme converges. A~thorough description of this algorithm is found in~\cite{GraphWalk}. 
\par
By projecting the voxels of the connected pore space onto the $x$~axis, we conclude whether each subsample has a~nontrivial permeability. Subsamples with zero permeability are excluded from the later workflow (the maximum flow value is exactly zero if and only if the permeability is zero---therefore no training on such data samples is necessary, cf.~\cref{SEC:MaxFlow}).
By encoding the simplified pore geometries in \mbox{1-bit} raw~format, memory consumption is \SI{125}{kB} per subsample. Accordingly, our whole library of data samples allocates only \SI{1.25}{GB} of disk space . Therefore, it is manageable on standard personal computers and file read/write access is reasonably cheap.
\par
We note that a~subsample of size $\SI{225}{\micro\meter}$ is too small to be a~representative elementary volume (REV) for most sandstone types, cf.~\cite{BentheimerPublication,PoreGeoLinxian}. As such, computed effective properties of one subsample cannot be expected to represent the whole segmented porous medium out of which it was extracted. On the other hand, the obtained training data set for our \mbox{PhyCNN} is therefore expected to be highly diverse, i.\,e.\ to contain highly permeable as well as narrow and confined pore geometries. Consequently, this setup is well suited to underline the robustness and prediction quality of our proposed methodology.
Moreover, hierarchical neural networks have proven to be a~powerful tool to leverage the performance of CNNs predicting permeability to samples of REV-size (in the order of $500^3$~voxels), cf.~\cite{Santos2021MultiscaleNetwork}. Therefore, we believe that our proposed methodology also benefits existing REV-scale models.

\subsection{Forward simulation}
\label{SEC:ForwardSim}
To apply a~supervised learning approach as outlined in~\cref{SEC:MachineLearningModel}, each subsample within the data set derived by the methods of~\cref{SEC:DataSampling} needs to be labeled with a~computed permeability $k_\mathrm{cmp}$ that we use as the reference value. 
\par
To this end, in~\cref{SEC:StokesSolve}, we perform flow simulations on the pore space. More precisely, for each of the $\num{10\,000}$ subsamples, a~stationary flow field along the $x$~direction is computed by solving the Stokes equations on the union of fluid voxels~$\Omega$ for the fluid velocity~$\vec{u}$ and pressure~$p$. The discretization uses arbitrary order (stable) Taylor--Hood or reduced-order stabilized Taylor--Hood mixed finite elements. In this paper, we consider voxel meshes only, yet the usage of unstructured grids by using obvious respective discrete spaces is straight forward.
\par
In~\cref{SEC:PermeabilityEstimation}, the (absolute, scalar) permeability is computed by averaging the pressure gradient and velocity field across the subsample in $x$~direction, cf.~\cref{SEC:DataSampling}. Apparently, as the data set contains $y$ and $z$-rotated versions of each subsample, this relates to the determination of the permeability with respect to all three main axes. By accurate bookkeeping, the diagonal permeability tensor of each initial subsample is retrievable. 
\par
Finally, in~\cref{SEC:DiscChoice}, we justify our choice of discretization parameters, i.\,e., mesh refinement level and finite element spaces used to produce the permeability values $k_\mathrm{cmp}$ to train and validate our \mbox{PhyCNN}. 
\par 
We note that---despite the finite resolution of the \muCT~images---the voxel scans are henceforth considered to represent the actual (ground-truth) pore geometry. Hence, errors arising from imaging are neglected as they pose an~independent part of the total methodology that is beyond the scope of this paper.

\begin{figure}[!ht]
\renewcommand\tabularxcolumn[1]{m{#1}} %
\begin{tabularx}{\linewidth}{@{}lCC@{}}
{} & \textbf{Subsample~0} & \textbf{Subsample~9213}\\
\rotatebox[origin=c]{90}{\small pressure} &
\includegraphics[width=\linewidth]{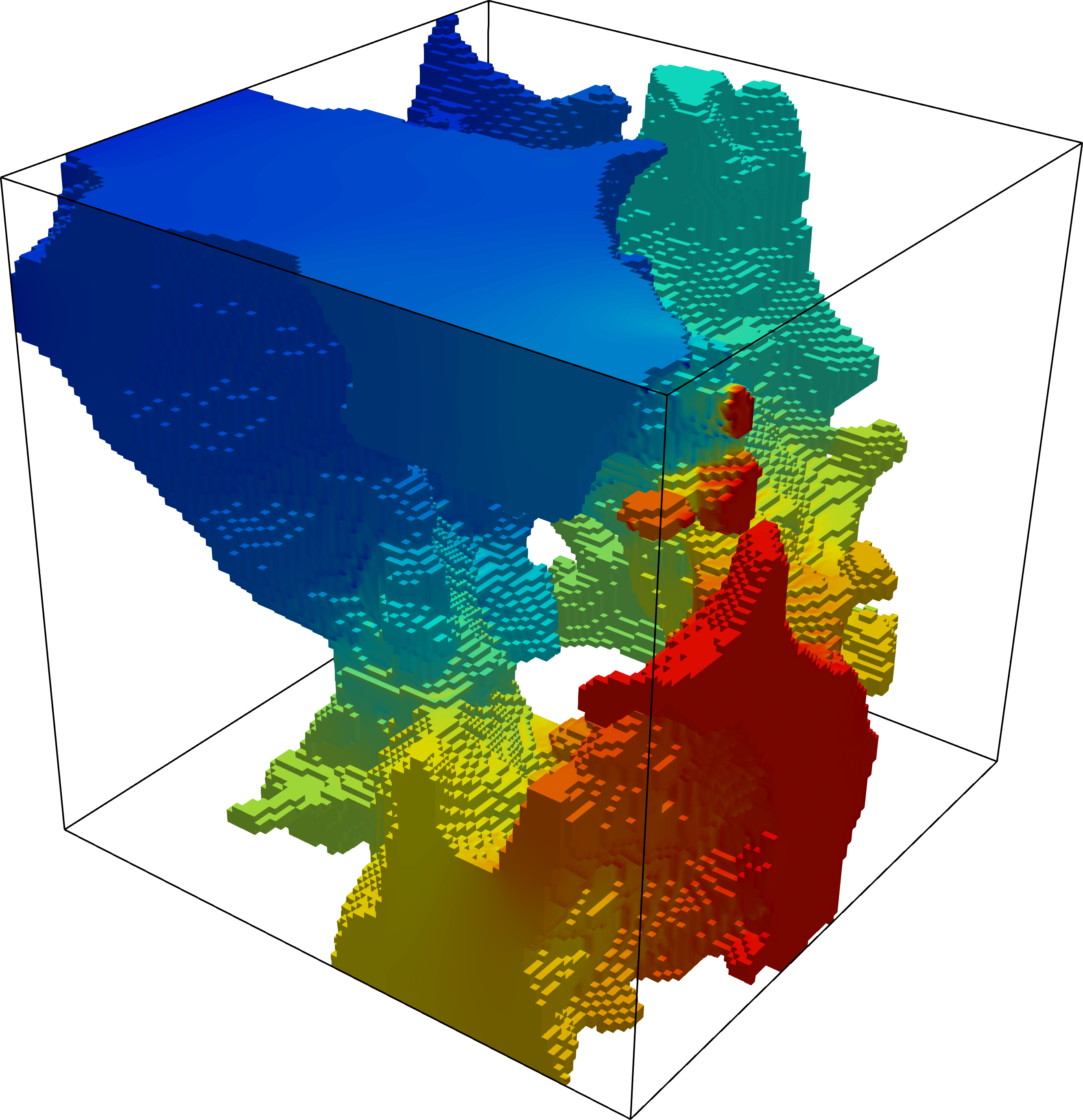}
& 
\includegraphics[width=\linewidth]{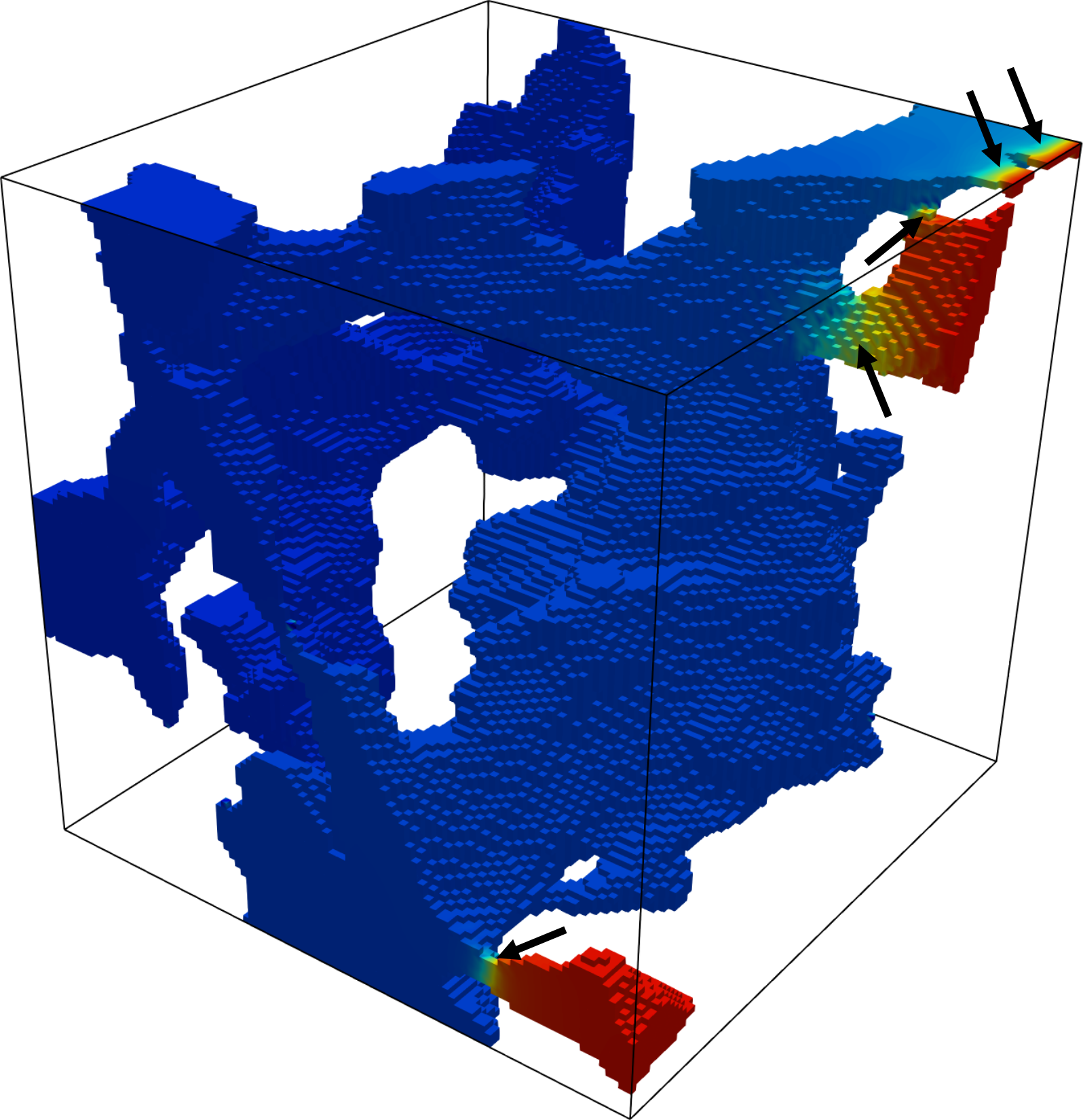}
\\
{} & \small $p_h\in[\num{-1.5E0},\num{1.4E0}]$ & \small $p_h\in[\num{-1.5E0},\num{1.8E0}]$
\\
\rotatebox[origin=c]{90}{\small velocity magnitude} &
\includegraphics[width=\linewidth]{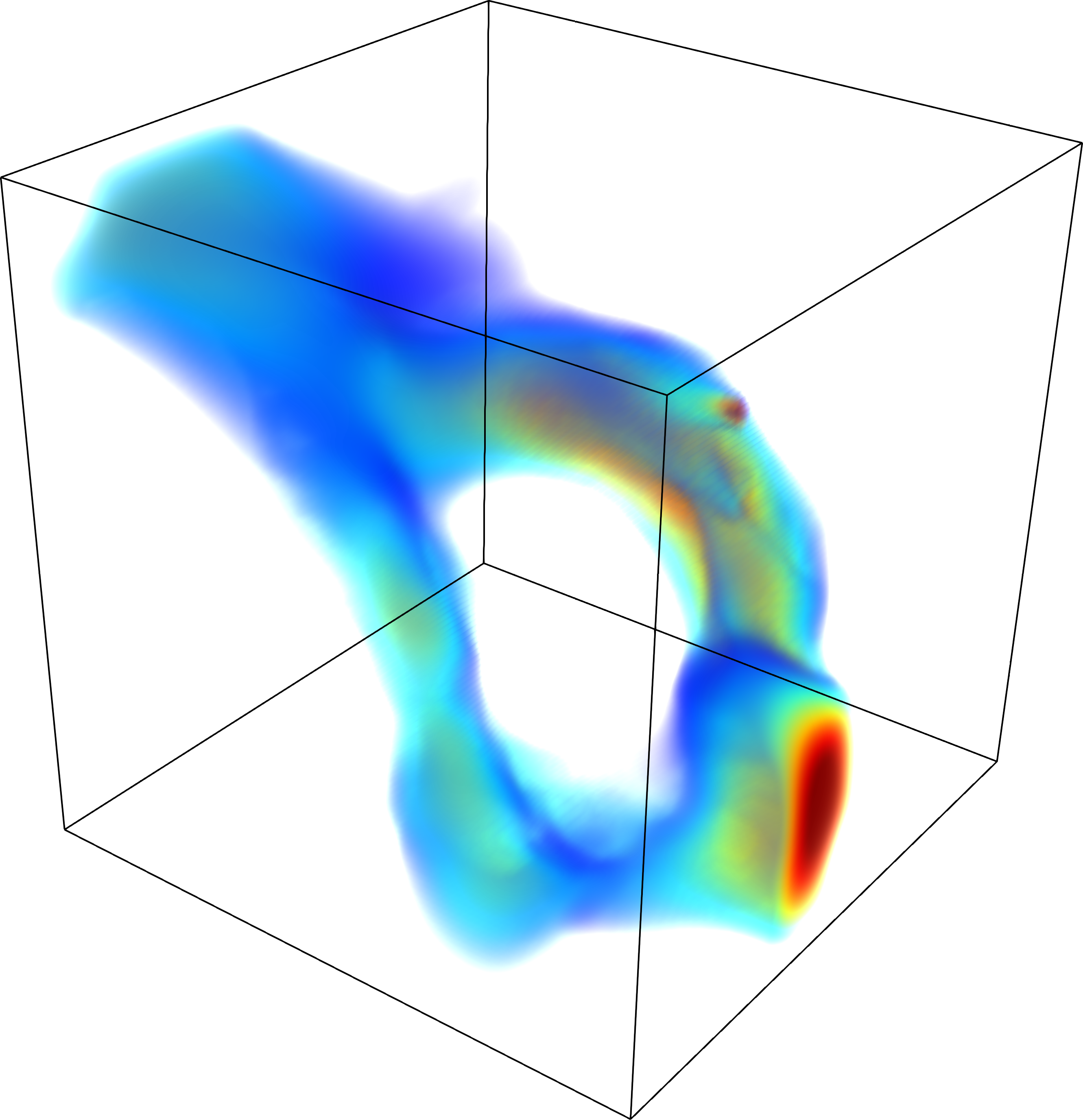}
&
\includegraphics[width=\linewidth]{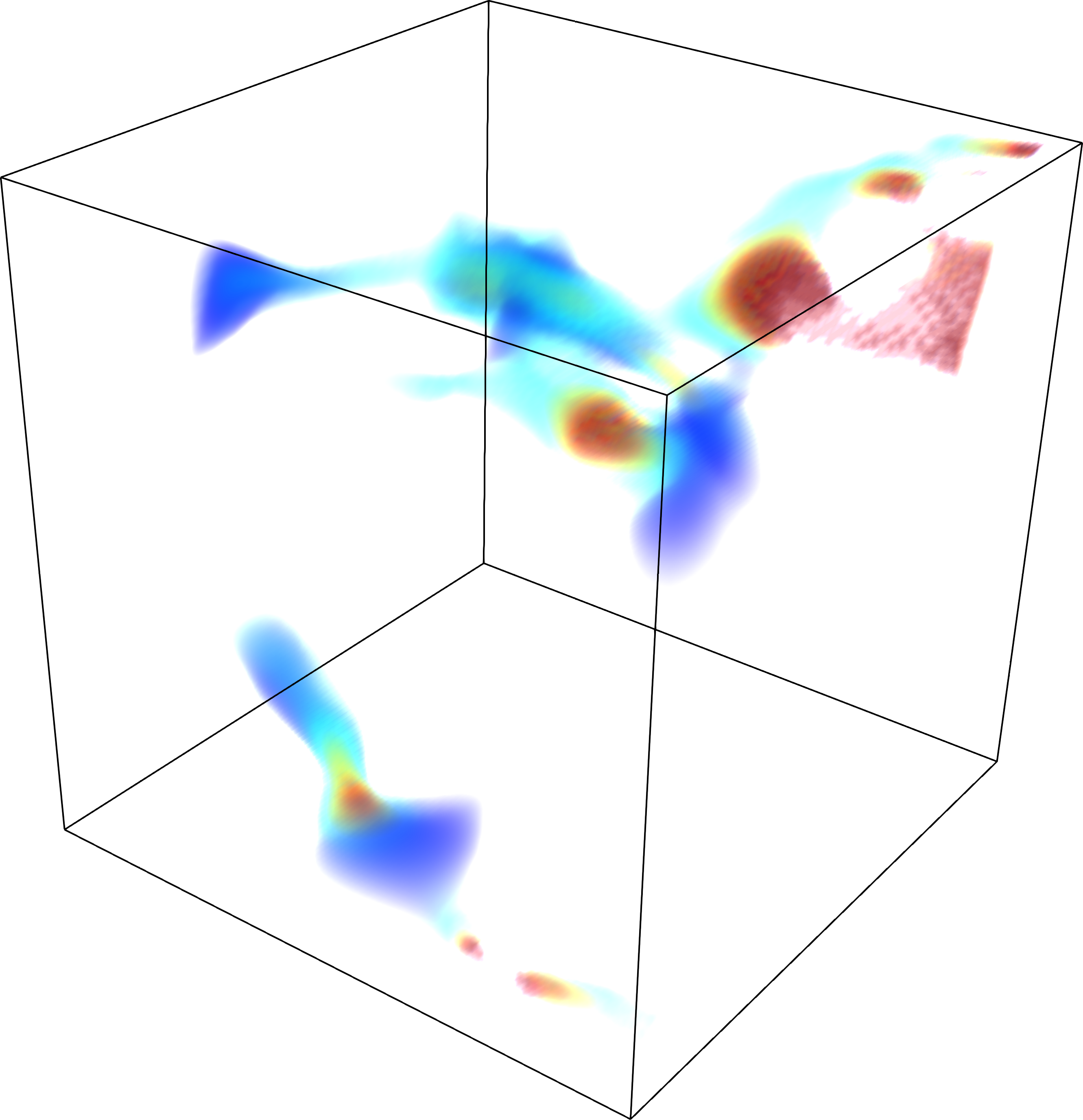}
\\
{} & \small $|\vec{u}_h|\in[0,\num{7.5E-3}]$ & \small $|\vec{u}_h|\in[0,\num{1.9E-3}]$
\end{tabularx}
\caption{Examples of geometries used for the training process, with pressure fields~\emph{(top)} and velocity magnitudes~\emph{(bottom)}. Subsample~0~\emph{(left)} exhibits moderate pressure gradients due to the wide and highly conductive channels in the pore space. Contrarily, in Subsample~9213~\emph{(right)} Stokes pressure is dominated by thin structures leading to an ill-conditioned problem, i.\,e., small changes in the diameter of narrow pores have drastic impact on the permeability. Thin pore throats are indicated by black arrows in the pressure plot.}
\label{FIG:CNNSamples}
\end{figure} 

\subsubsection{Computation of the flow field}
\label{SEC:StokesSolve}
We consider the stationary Stokes equation for a~Newtonian fluid in the nondimensionalized form,
\begin{subequations}
\label{EQ:StokesProblem} 
\begin{align}
  -\frac{1}{\Re}\laplace\vec{u} + \grad p                &= \vec{0}\quad \text{in}~\Omega,\label{EQ:StokesProblem:MomentumConservation}\\
  \div\vec{u}                               &= 0\quad \text{in}~\Omega,\label{EQ:StokesProblem:MassConservation}
\end{align}
where $\Omega\subset(0,1)^3$ is a~domain that consists of the union of fluid voxels of a~considered $100{\times}100{\times}100$ voxel subsample (i.\,e.\ the subsample is inscribed into the unit cube).
In~\eqref{EQ:StokesProblem}, $\vec{u}=\vec{u}(x,y,z)$ denotes the (dimensionless) fluid velocity, $p=p(x,y,z)$~the (dimensionless) pressure and $\Re\coloneqq \rho\, U_\mathrm{c} L_\mathrm{c}/\mu$ the Reynolds number of the system with characteristic length~$L_\mathrm{c}$~\si{[\meter]}, characteristic velocity~$U_\mathrm{c}$~\si{[m.s^{-1}]}, fluid density~$\rho$~\si{[kg.m^{-3}]}, and fluid viscosity~$\mu$~\si{[\pascal.\second]}. Note that the permeability~$k_\mathrm{cmp}~\si{[m^2]}$ is invariant with respect to~$\Re$ (see below).
The data set is constructed in such a~way that there are connected fluid voxels (i.\,e.~sharing a~common face) reaching from the~$x{=}0$~plane to the $x{=}1$~plane,  cf.~\cref{FIG:CNNSamples}, since impermeable subsamples were excluded from the workflow in the \enquote*{data acquisition} step, cf.~\cref{fig:FlowChart} and~\cref{SEC:DataSampling}. 
The flow field is driven by a~pressure gradient in $x$~direction induced by the boundary condition
\begin{equation}
  \left(\frac{1}{\Re}\grad\vec{u} - p\,\vecc{I}\right)\normal = \vec{e}_x\quad \text{on}~\Gamma_{\mathrm{N}},\label{EQ:StokesProblem:NeumannBC}
\end{equation}
where $\Gamma_{\mathrm{N}} \coloneqq \big\{(x,y,z)\in\partial\Omega \,|\, x\in\{0,1\}\big\}$ and $\vec{e}_x$ denoting the unit vector in $x$~direction. On the remaining boundary $\Gamma_\mathrm{D}\coloneqq \partial\Omega\setminus\Gamma_\mathrm{N}$, no-slip boundary conditions are prescribed,
\begin{equation}
\vec{u} = \vec{0}\quad \text{on}~\Gamma_\mathrm{D}.\label{EQ:StokesProblem:DirichletBC}
\end{equation}
\end{subequations}
\par
The weak formulation of~\eqref{EQ:StokesProblem} is discretized by generalized Taylor--Hood pairs of spaces, $\IQ_{\ell+1}^3/\IQ_\ell$, where $\IQ_\ell$ denotes the local space of polynomials of degree at most~$\ell$ in each variable~$x,y,z$, cf.~\cite{ElmanEtAl2014Book,ErnGuermond2004Book}. For~\mbox{$\ell=0$}, the pressure space~$\IQ_0=\IP_0$ is discontinuous and consists of elementwise constants. The respective \enquote{reduced Taylor--Hood pair}~$\IQ_1^3/\IP_0$ requires stabilization (see below) due to a~lack of discrete inf-sup-stability~\cite{ElmanEtAl2014Book}.
Let $\vec{\phi}_i = \vec{\phi}_i(x,y,z):\Omega\rightarrow\IR^3$, $i=1,\ldots,\numU$ and $\psi_i = \psi_i(x,y,z):\Omega\rightarrow\IR$, $i=1,\ldots,\numP$ denote the basis functions for the global discrete spaces for velocity and pressure, respectively.
The discrete velocity~$\vec{u}_h=\vec{u}_h(x,y,z)$ and discrete pressure~$p_h=p_h(x,y,z)$ then have the representation
\begin{equation}\label{EQ:SolutionRepresentation}
  \vec{u}_h=\sum_{i=1}^{\numU}[\xU]_i\,\vec{\phi}_i,\qquad
  p_h = \sum_{i=1}^{\numP}[\xP]_i\,\psi_i
\end{equation}
with degree-of-freedom vectors $\xU\in\IR^{\numU}$, $\xP\in\IR^{\numP}$, which are unique solutions of the linear system
\begin{align}
  \label{EQ:DiscretizedProblem}
  \begin{bmatrix}
  \vecc{A} & \phantom{-}\vecc{B}^\tp \\
  \vecc{B} & -\vecc{C}
  \end{bmatrix}
  \begin{bmatrix}
  \xU\\
  \xP
  \end{bmatrix}=
  \begin{bmatrix}
  \bU\\
  \vec{0}
  \end{bmatrix}
  \qquad\Longleftrightarrow:\qquad
  \mathcal{A} \, \vec{x} = \vec{b}
\end{align}
with right-hand side $\bU\in\IR^{\numU}$, $[\bU]_i \coloneqq  \int_{\Gamma_\mathrm{N}}\vec{e}_x\cdot\vec{\phi}_i$.
The sparse blocks in~$\mathcal{A}$ are the vector-Laplacian matrix~$\vecc{A}\in\IR^{\numU, \numU}$ and the divergence matrix~$\vecc{B}\in\IR^{\numP, \numU}$,
\begin{equation*}
  [\vecc{A}]_{i,j}  \coloneqq  \int_\Omega \grad\vec{\phi}_i : \grad\vec{\phi}_j,
  \qquad
  [\vecc{B}]_{k,j}  \coloneqq - \int_\Omega \psi_k\div\vec{\phi}_j,
\end{equation*}
and $\vecc{C}$ is a~stabilization matrix that is required only for lowest order~$\ell=0$ to guarantee the full rank of~$\mathcal{A}$. We choose $\vecc{C}$ as in~\cite{ElmanEtAl2014Book},~(3.84), in which case, $\vecc{C}$
can be interpreted as a~pressure-Laplacian discretized by cell-centered finite volumes~\cite{FALR2017AdvCH}.
For $\ell>0$, $\vecc{C}$ is set to zero. In either case, $\mathcal{A}$ is symmetric and indefinite with $n$~positive and $m$~negative eigenvalues.
\par
In order to solve the saddle point system~\eqref{EQ:DiscretizedProblem} efficiently, a~preconditioned MINRES method is applied, as it is the best choice of Krylov subspace methods for symmetric indefinite systems~\cite{Wathen2015Preconditioning}. The chosen precondition operator for~$\mathcal{A}$ in~\eqref{EQ:DiscretizedProblem} is the symmetric and positively definite block-diagonal matrix 
\begin{equation*}
  \mathcal{P}  \coloneqq \mathrm{diag}\,(\vecc{A}, \vecc{W})
\end{equation*}
with $\vecc{W}$ being the pressure-mass matrix $[\vecc{W}_{k,l}]\coloneqq\int_\Omega \psi_k\,\psi_l$. The action of the inverse~$\mathcal{P}^{-1}$ in each Krylov iteration is approximated block-wise by one V-cycle of an~algebraic multigrid method (Boomer AMG from the Hypre library~\cite{HYPRE}). Since~$\vecc{W}$ is spectrally equivalent to the (negative) Schur complement~$\vecc{B}\vecc{A}^{-1}\vecc{B}^\tp$ of~$\mathcal{A}$ (for $\ell>0$), and due to the utilization of AMG, the number of Krylov iterations required to reach a~given relative tolerance is bounded independently of the mesh size~\cite{Benzi2005Saddle,Rozloznik2018Saddle,PearsonPestanaSilvester2017} (however, it highly depends on the geometry of the domain). MINRES with preconditioning as described above belongs to the state-of-the art Stokes solvers in the high-performance computing context~\cite{GmeinerRuedeEtAl2016}.
In~\cref{SEC:DiscChoice}, we discuss appropriate choices regarding mesh size and discretization order.  
\par
\cref{FIG:CNNSamples}~illustrates Stokes velocity and pressure fields for two exemplary subsamples exhibiting qualitatively highly different pore-space geometries such as wide pore throats and narrow channels. We will emphasize the impact of highly and merely permeable rock samples on the behavior of DNS-based permeability computations and \mbox{PhyCNN} predictions throughout the paper.

\subsubsection{Permeability estimation}
\label{SEC:PermeabilityEstimation}
We deduce the permeability value $k_\mathrm{cmp}$ of interest from the previously calculated Stokes velocity $\vec{u}$ and pressure $p$ (we suppress the discretization index~$h$ in this section). In the Stokes~equations~\eqref{EQ:StokesProblem}, the inflow and outflow boundaries are defined as
\begin{equation*}
\Gamma_{\mathrm{in}} \coloneqq  \big\{(x,y,z)\in\partial\Omega \,|\, x = 0 \big\}\,,
\qquad 
\Gamma_{\mathrm{out}} \coloneqq \ \big\{(x,y,z)\in\partial\Omega \,|\, x = 1 \big\}
\end{equation*}
and thus disjoint subsets of~$\Gamma_\mathrm{N}$.
From the solution~$(\vec{u},p)$ of~\eqref{EQ:StokesProblem}, we compute the approximated permeability~$k_\mathrm{cmp}$~\si{[m^2]} by the classical Darcy law,
which reads in nondimensionalized form,
\begin{subequations}\label{EQ:PermEstFormula}
\begin{equation}\label{EQ:PermEstFormula:a}
Q = -\Da\,\Re\,(P_\mathrm{out}-P_\mathrm{in}),
\end{equation}
with (dimensionless) volume flow rate $Q$, and (dimensionless) area-averaged inflow and outflow pressures, $P_\mathrm{in}$, $P_\mathrm{out}$, given by~\cite{LiuFAR2019Permeability}
\begin{equation*}
Q \coloneqq \int\limits_{\Gamma_\mathrm{out}}\vec{u}\cdot\normal 
\,,\qquad
P_\mathrm{in} \coloneqq \frac{1}{\vert \Gamma_\mathrm{in}\vert}\int\limits_{\Gamma_\mathrm{in}}p
\,,\qquad
P_\mathrm{out} \coloneqq \frac{1}{\vert \Gamma_\mathrm{out}\vert}\int\limits_{\Gamma_\mathrm{out}}p\, .
\end{equation*}
From \eqref{EQ:PermEstFormula:a}, $k_\mathrm{cmp}$ is derived from the Darcy number
\begin{equation}\label{EQ:PermEstFormula:b}
\Da \coloneqq \frac{k_\mathrm{cmp}}{L_\mathrm{c}^2}\,,
\end{equation}
\end{subequations}
where~$L_\mathrm{c}$ is the characteristic length, in our case, the edge length of a~$100^3$-voxel subsample, i.\,e., $L_\mathrm{c}=\SI{225}{\micro\meter}$. 
We choose~$\Re$ equal to one, since $\Re$ does not influence the permeability value~$k_\mathrm{cmp}$ (a~rescaling of~$\vec{u}$ by~$\Re^{-1}$ implies a~rescaling of~$Q$ by~$\Re^{-1}$ and therefore, $\Re$ cancels out in~\eqref{EQ:PermEstFormula:a}).
\par
Formula~\eqref{EQ:PermEstFormula:a} determines an approximation of the \emph{scalar} permeability in $x$~direction by area averaging.  
If this approach was applied to all three principal directions, it yielded a~diagonal permeability tensor.
In~\cite{Guibert2015}, a~volume-averaged approach is proposed that is capable of determining the full permeability tensor. Application to $x$~direction only, yields a~column vector, whose entries are in general non-trivial. Since we want to train our \mbox{PhyCNN} with one scalar permeability value only, we decided to utilize the area-averaging approach in this study.

\subsubsection{Choice of discretization parameters}
\label{SEC:DiscChoice}

In this section, we investigate the influence of mesh refinement and choice of polynomial order on the solution quality by comparing the computed permeability $k_\mathrm{cmp}$ obtained from~\eqref{EQ:PermEstFormula:a} to the analytical value $k_\mathrm{ana}$. In order to have analytical results available, we restrict our considerations to viscous flow through rectangular channels~$\Omega_{a, b}$. This is supposed to constitute a~sufficient benchmark, since the approximation quality of the computed permeability~$k_\mathrm{cmp}$ is dominated by the discretization error in narrow pore throats due to high local gradients. 
\par
For the Stokes equations~\eqref{EQ:StokesProblem},
consider a~rectangular channel 
\begin{equation*}
\Omega_{a, b}\coloneqq (0,1)\times \left(\frac{1}{2}-\frac{a}{2},\frac{1}{2}+\frac{a}{2}\right)\times \left(\frac{1}{2}-\frac{b}{2},\frac{1}{2}+\frac{b}{2}\right)\subset\IR^3
\end{equation*} 
of width $a\in \left(0,\frac{1}{2} \right)$ and height $b\in \left(0,\frac{1}{2} \right)$.
An~analytical expression for its permeability~$k_\mathrm{ana}$ is~\cite{bruus2008theoretical}
\begin{align*}
k_\mathrm{ana} &\coloneqq \frac{K_\infty\cdot\min(a,b)^3\cdot\max(a,b)}{12},\qquad\text{where} \\ 
    K_j&\coloneqq 1-\sum_{n=1}^{j}\frac{1}{(2n-1)^5}\cdot\frac{192}{\pi^5}\cdot\frac{\min(a,b)}{\max(a,b)}\tanh\left((2n-1)\frac{\pi}{2}\frac{\max(a,b)}{\min(a,b)}\right),
\end{align*}
denoting the limit of $K_j$ for $j\to \infty$ by $K_\infty$. By application of the triangular inequality, \mbox{$0<\min(a,b)\leq \max(a,b)$}, and $|\tanh(x)|<1\;\forall x\in\IR$, we obtain the following approximation error bound:
\begin{align*}
    |K_\infty-K_j|\leq \sum\limits_{n=j+1}^\infty \frac{1}{(2n-1)^5}\cdot\frac{192}{\pi^5}. 
\end{align*}
Piecewise application of Jensen's inequality to the convex function $(2x-1)^{-5}$ finally yields the estimate:
\begin{align*}
     |K_\infty-K_j|\leq  \frac{192}{\pi^5}  \int\limits_{j+0.5}^{\infty} \frac{1}{(2x-1)^5} \;\dd x = \frac{192}{128} \cdot \frac{1}{\pi^5 j^4} \approx 0.0049 \,j^{-4}.
\end{align*}
As such, we obtain at least six significant digits using $K_{10}$ for the calculation of the analytical reference permeability~$k_\mathrm{ana}$.
\par
\begin{table}[ht!]
  \centering
  \begin{tabular}{ c c c c r r}
   \toprule
    order~$\ell$ & ref.~level  & $k_\mathrm{cmp}$ & $\dfrac{\vert k_\mathrm{cmp}-k_\mathrm{ana}\vert}{k_\mathrm{ana}}$ & $m$ (DOF~$\vec{u}_h$) & $n$ (DOF~$p_h$) \\
    \midrule
    0  & 0 & \num{8.44E-8} & \num{8.82E-2} & $8\,484$ & $1\,800$ \\
    0  & 1 & \num{9.06E-8} & \num{2.22E-2} & $54\,873$ & $14\,400$\\
    0  & 2 & \num{9.22E-8} & \num{4.62E-3} & $390\,975$ & $115\,200$ \\
    1  & 0 & \num{9.25E-8} & \num{1.20E-3} & $54\,873$ & $2\,828$ \\
    1  & 1 & \num{9.26E-8} & \num{4.06E-4} & $390\,975$ & $18\,291$ \\
    \bottomrule
  \end{tabular}
  \caption{Computed permeabilities~$k_\mathrm{cmp}$ of a~channel with~$6{\times}3$ voxel rectangular cross section for different polynomial orders~$\ell$ of the pressure space (cf.~\cref{SEC:StokesSolve}) and mesh refinement levels. As in~\eqref{EQ:SolutionRepresentation}, $m$ and $n$ are the numbers of DOF for velocity~$\vec{u}_h$ and pressure~$p_h$, respectively.  The relative tolerance of MINRES is set to \num{1.0E-6}, which has shown to be sufficient for three significant digits in the permeability value using a~small scouting test set. }
  \label{TAB:PermAnaTest}  
\end{table}

\cref{TAB:PermAnaTest}~lists the relative error for the computed permeability~$k_\mathrm{cmp}$ obtained from \eqref{EQ:PermEstFormula:a} on~$\Omega_{0.06,0.03}$ using different polynomial orders~$\ell$ and global mesh refinement levels. As expected, both refinement and higher-order discrete spaces consistently reduced the error with respect to the analytical solution. The best cost-to-approximation-quality ratio is achieved by $\IQ_2^3/\IQ_1$~elements ($\ell=1$) on the original grid. In particular, this choice poses a~significant improvement in accuracy over the use of lowest-order spaces~$\IQ_1^3/\IQ_0$ ($\ell=0$) with two global mesh refinement levels.
\par
The advantage of higher order discretizations presented above for narrow rectangular channels seems to generalize to our actual geological samples, which include confined structures. For Subsample~0, cf.~\cref{FIG:CNNSamples}, where the main flow channels are wide with respect to the voxel resolution, $k_\mathrm{cmp}$ is hardly affected by increasing the polynomial order~$p$ from zero to one ($4\%$ deviation). On the other hand, Subsample~9213 (\cref{FIG:CNNSamples}) exhibiting narrow and branchy structures experienced a~significant relative increase in computed permeability of $43\%$. Therefore, we approve the increased computational complexity of lowest order classical Taylor--Hood elements ($\ell=1$) for permeability labeling in our forward simulations, cf.~\cref{SEC:StokesSolve}. We emphasize that our scheme does not require mesh refinement, which is typically required in LBM simulations of natural rock.
\par
This completes the methodology description for data acquisition and yields a~training set suitable to train our neural network for permeability prediction. We conclusively note that the solver converged for every subsample in the database particularly maintaining the generality of our training set. 

\subsection{Evaluation metrics}
\label{SEC:EvaluationMetrics}
In our study, we use the following well-known metrics to characterize and compare different measures of a~data set statistically.
\par
First, we define the standard deviation $\sigma$ by
\begin{align*}
    \sigma = \sqrt{\frac{1}{N-1} \sum\limits_{i=1}^N \left(t_i-\bar{t} \right)^2}
\end{align*}
for a~number $N$ of real-valued data $t_i$ and arithmetic mean value $\bar{t}$, measuring the expected deviation from~$\bar{t}$. 
In order to quantify the correlation between different measures of a~data set, we further introduce the coefficient of determination $R^2$
\begin{align*}
    R^2 = 1-\frac{\sum\limits_{i=1}^N (t_i-y_i)^2}{\sum\limits_{i=1}^N (t_i-\bar{t})^2}
\end{align*}
for targets $t_i$ and corresponding predictions $y_i$,  encoding the share of variance in the targets that is covered by the predictions.

Finally, we define the mean-squared error MSE by
\begin{align*}
\text{MSE}= \sum\limits_{i=1}^N \frac{(t_i-y_i)^2}{N}.   
\end{align*}

\section{Machine learning model}
\label{SEC:MachineLearningModel}
In this section, we motivate and present our suggested neural network architecture for permeability prediction. Therefore, we provide a~brief introduction to the class of convolutional neural networks in~\cref{SEC:CNN} as the basis for our specific network architecture. For further reading, we refer to~\cite{aggarwal2018neural}. In~\cref{SEC:phyCNN}, this model is augmented by an additional physical input quantity, which is derived in~\cref{SEC:MaxFlow} in detail. For simplicity, we refer to the $100^3$-voxel geological subsamples derived in~\cref{SEC:DataSampling} as (data) samples in the context of supervised machine learning. 

\subsection{Convolutional neural networks}
\label{SEC:CNN}
In the following, we discuss the building blocks required to set up an artificial neural network tailored to the task of permeability predictions from 3D~binary images of pore-space geometries. In a~broader sense, we start with the more general concept of feed-forward neural networks.
\par
As indicated by the name, feed-forward neural networks (FFNs) process their input data by successively propagating the input information through a~finite number of layers. Each of these layers $i\in \{0,\dots,L \}$ consists of $N_i$ neurons, where the indices $0$ and $L$ correspond to the network's input and output layer, respectively. Typically, the action of a~layer $1\leq i \leq L$ on input data~$x_{i-1}\in\IR^{N_{i-1}}$ is given as an~affine-linear transformation, followed by application of a~nonlinearity (activation function) $\sigma_i$:
\begin{align*}
    x_i = \sigma_i \left(W_i  x_{i-1} +b_i\right).
\end{align*}
We refer to $W_i\in \IR^{N_{i},N_{i-1}}$ as the weight matrices and to $b_i\in \IR^{N_{i}}$ as the biases. Their combined entries are the learnable parameters of the network.
\par
Convolutional neural networks (CNNs) constitute a~subclass of FFNs exhibiting a~dedicated internal structure specialized to the interpretation of image data. Furthermore, CNNs naturally incorporate translational invariance with respect to the absolute spatial position of features within the input data. In the following, we give a~short introduction to this kind of networks. For a~broader overview of common network designs and layer types, we refer to the respective literature, e.\,g.~\cite{aggarwal2018neural}. A~list of the different layer types used in our study is given in~\cref{TAB:CNNlayout}.
\par
Typically, CNNs consist of several convolutional (\texttt{conv}) lower layers (filter layers) including pooling operations, followed by \texttt{dense} neuron layers. As indicated by the name, convolutional layers perform the discrete analog of a~convolution, where the convolution kernel is a~learnable parameter of the network. As such, they possess a~highly structured weight matrix~$W$ and are therefore superior to fully connected layers in terms of computational complexity, especially for highly localized kernels (sparsity). This restriction is justified in our application scenario, since most (low-level) geometry features are based on highly localized correlations among neighboring voxels. 
\par
MaxPooling (\texttt{MP}) layers are used to reduce the resolution of the output image. Presenting a~natural bottleneck, the most significant pieces of information are selected and passed to the subsequent neuron layers. Application of several different filters to the same input within each layer of the network extracts increasingly complex characteristic features of the pore geometry. Finally, this information is interpreted by the subsequent \texttt{dense} neural layers.
\par 
Convergence of the training procedure is improved by exploiting additional batch normalization (\texttt{BN}) layers, calibrating the mean and spread of the provided inputs. Throughout the network, we apply \texttt{LeakyReLU} nonlinearities in an elementwise manner. For a~parameter~$\alpha\in (0,1)$, the \mbox{\texttt{LeakyReLU}($\alpha$)} function is defined by
\begin{align*}
    \texttt{LeakyReLU}(\alpha; x) \coloneqq \max(0,x)-\alpha \max(0,-x). 
\end{align*}
Due to the slope of this nonlinearity being bounded away from zero, we circumvent the \textit{dying neuron problem} occurring in deep neural networks with \texttt{ReLU} nonlinearities~\cite{dyingRELU}, which arise formally by setting $\alpha=0$. Furthermore, backpropagation during the learning phase is facilitated. Our implementation is conducted using the Deep Learning Toolbox in Matlab~R2021a~\cite{MATLAB:2021a}.

\subsection{Physics-informed convolutional neural networks}
\label{SEC:phyCNN}
In the following, we use the building blocks for general CNNs as introduced in~\cref{SEC:CNN} and derive a~network architecture specifically tailored to our application in permeability prediction. Therefore, we first give an introduction to \mbox{PhyCNNs} and thereafter discuss the choice of meaningful additional input quantities to the network. 

\subsubsection{General architecture}
\begin{figure}[!ht]
    \centering
    \includegraphics[width=\textwidth]{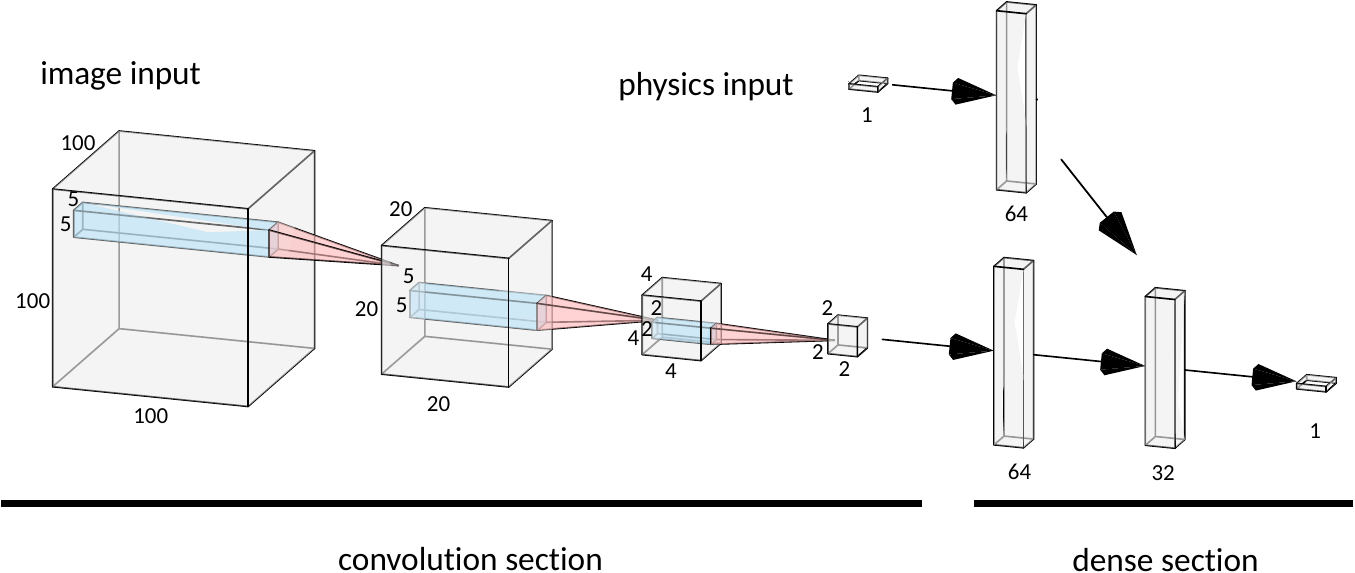}
    \caption{Schematic representation of physics-informed convolutional neural networks \mbox{(PhyCNNs)}. Numbers at the layers denote their dimension (number of neurons). Note that the cubic convolutional kernels and their dimension are illustrated in 2D for clarity. The scalar valued physical input circumvents the image convolution layers and is directly fed into the subsequent \texttt{dense} layers. Graphics produced using \mbox{NN-SVG}~\cite{ANNDrawerLeNail2019}.}
    \label{FIG:PhyCNNStructure}
\end{figure}

As discussed above, standard CNNs are required to extract all important features and information solely from the input image data to predict the correct output. Nevertheless, it is possible to provide the network with additional input derived from the image data in a~preprocessing step, leading to the field of \mbox{PhyCNNs}. An~exemplary schematic representation of such a~network structure is provided in~\cref{FIG:PhyCNNStructure}. As the additional physical input is typically not of image type (and hence does not exhibit spatially correlated features), it omits the convolutional layers and is directly coupled to the upper \texttt{dense} layers. As such, \mbox{PhyCNNs} predict by considering both the extracted geometry features as well as the provided physical input quantities. Depending on the quality and quantity of the extracted features as well as the specific application, these alone may even provide sufficient information to the network to accurately perform predictions, cf.~\cite{ArayaPoloEtAl2018}, underlining the potential of physics-informed strategies.
\par
This approach has been successfully applied to the prediction of permeabilities from pore-space geometries: In~\cite{SantosEtAl2020}, the Euclidean distance map, maximum inscribed sphere radii, and \emph{time of flight} maps are inserted as inputs to determine the overall fluid field. In~\cite{CNNPerm}, porosity and surface area have been used to increase predictions performance for 2D image data. Similar observations have been made on 3D data where the additional consideration of porosity and tortuosity significantly reduced the number of outliers, especially when training on small data sets, cf.~\cite{TANGPhyCNN}. The specific choice of these parameters is related to the Kozeny--Carman formula estimating permeability from porosity and surface area / tortuosity. As indicated in~\cite{SchulzRayZech}, the proposed relation deteriorates in quality for increasingly complex geometries. However,~\cite{CNNTEMBELY} found porosity to be the most prominent input factor for their CNN among a~set of various derived quantities such as coordinate number or mean pore size. Furthermore, characteristics obtained by pore network modeling have shown low general accuracy~\cite{CNNTEMBELY}. 
\par
\begin{figure}[!ht]
    \centering
    \includegraphics[width=\linewidth]{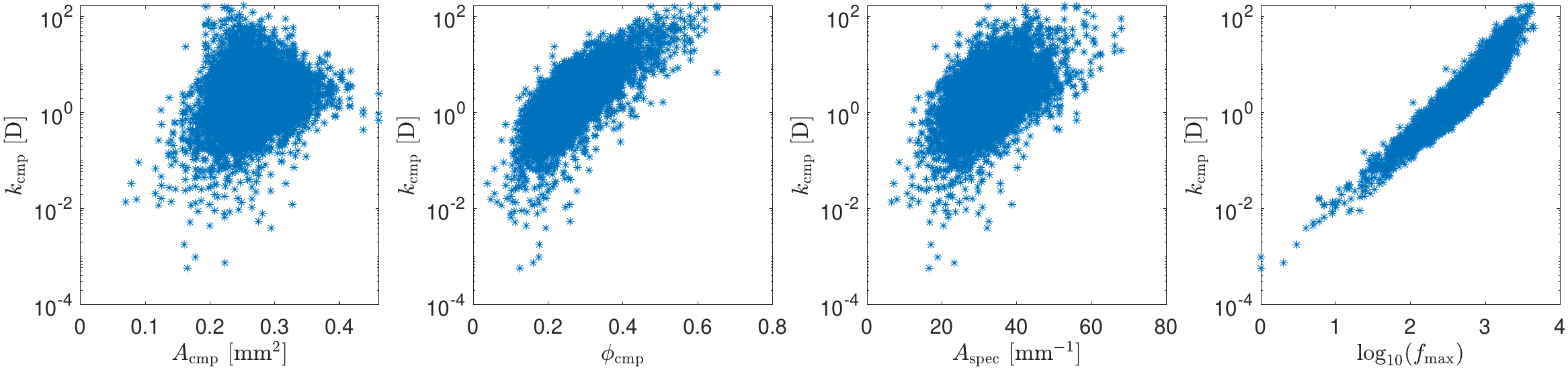}
    \caption{Correlation of different physical measures of the underlying geometry with the computed permeability values $k_\mathrm{cmp}$ for Bentheimer data samples. From left to right: Computed surface area $A_\mathrm{cmp}$ ($R^2=0.055$), computed porosity $\phi_\mathrm{cmp}$ ($R^2=0.547$), specific surface area $A_\mathrm{spec}$ ($R^2=0.455$), maximum flow on graph representation ${f_\mathrm{max}}$ ($R^2=0.869$). The given $R^2$~values refer to a linear regression model.}
    \label{FIG:PhysicsCorrelation}
\end{figure}

As indicated by our data set statistics in~\cref{FIG:PhysicsCorrelation}, porosity and surface area are only weakly correlated with the computed permeability $k_\mathrm{cmp}$ for our complex 3D~setting. For a~given surface area $A_\mathrm{cmp}$, we observe data samples covering four orders of magnitude in their respective permeability values, about half the range is observed for fixed porosities $\phi_\mathrm{cmp}$. Hence, drastic increases in accuracy cannot be expected by regarding these quantities as additional physical input in our case. Instead, we construct another more specific physical quantity that can be computed efficiently from the input data using graph algorithms included in Matlab~\cite{MATLAB:2021a}. More precisely, we solve a~surrogate maximum flow problem on a~graph representation of the pore space. In~\cref{SEC:MaxFlow}, we give a~thorough description of this chosen quantity. 

Our final network architecture is presented in~\cref{TAB:CNNlayout} and~\cref{FIG:PhyCNNStructure}. The structure is based on the findings of~\cite{RapidEstimate}, where the hyper-parameters of a~purely convolutional neural network have been optimized using a~grid search algorithm. For our study, we decrease the bottleneck constituted by the final convolutional layer and improve convergence by adding batch normalization layers and relaxed cut-offs as nonlinearities (\texttt{LeakyReLU}). Furthermore, the maximum flow value~${f_\mathrm{max}}$ is treated as an additional scalar physical input quantity which is duplicated to a~vector of length~$64$ and subsequently concatenated with the first \texttt{dense} layer of the network. By doing so, we balance both inputs to the second \texttt{dense} layer which results in a~more uniform weight distribution. 

\begin{table}[!ht]
    \centering
    \begin{tabular}{l l r}
      \toprule
      \textbf{block}  & \textbf{layers} & \textbf{learnables}                               \\
      \midrule
      input1  & image input 100{$\times$}100{$\times$}100 & ---               \\
      input2  & physics input 1{$\times$}1{$\times$}1 &  ---                \\
      conv1  & \texttt{conv}(32,5) -- \texttt{BN} -- \texttt{LeakyReLU}(0.1)  -- \texttt{MP}(5,5) & 4\,096 \\
      conv2  & \texttt{conv}(64,5) -- \texttt{BN} -- \texttt{LeakyReLU}(0.1) -- \texttt{MP}(5,5) & 256\,192 \\
      conv3  & \texttt{conv}(128,3) -- \texttt{BN} -- \texttt{LeakyReLU}(0.1) -- \texttt{MP}(2,2) & 221\,568  \\
      dense1 & \texttt{dense}(64) -- \texttt{LeakyReLU}(0.1)   & 65\,728              \\
      dense2 & \texttt{dense}(32) -- \texttt{LeakyReLU}(0.1)   & 4\,161              \\
      output & \texttt{regression}(1) &  ---  
      \\\bottomrule
      \vspace{0.05cm}
    \end{tabular}
    \caption{Layer structure of our \mbox{PhyCNN}. The physical scalar input2 is expanded to dimension $64{\times}1$ and concatenated with the output of \enquote*{dense1} using a~depth concatenation layer. Nomenclature:
    \\
    \hspace*{0.1\linewidth}\begin{minipage}[t]{0.9\linewidth}
        \begin{tabbing}
          \texttt{conv}($N$,$K$): \hspace{1em} \=  convolutional layer with $N$ channels and $K\times K$ kernel size;\\
          \texttt{BN}: \> batch normalization layer;\\
          \texttt{MP}($N$,$P$):     \>maxPooling Layer, size $N$ stride $P$;\\
          \texttt{dense}($N$):   \>dense layer with $N$ neurons;\\
          \texttt{regression}($N$): \>regression layer with $N$ neurons;\\
          \texttt{LeakyReLU}($\alpha$): \>leaky rectified linear unit, slope $\alpha$ on negative inputs.
        \end{tabbing}
      \end{minipage}
    } %
    \label{TAB:CNNlayout}
  \end{table}

\subsubsection{Maximum flow problems on graphs}
\label{SEC:MaxFlow}
In the field of optimization, maximum flow problems aim at identifying the maximum flow rate a~network of pipes is able to sustain. Let the network be described by an undirected graph $G(N,E,\omega)$, where the nodes~$N$ correspond to distribution nodes, edges~$E$ to connections by pipes, and the weights $\omega$ encode the maximal capacity of each pipe. 
For any two nodes $n_1\neq n_2 \in N$ (source and sink), we can determine the maximum flow between those nodes allowed by the network. For a~thorough introduction to graph algorithms and maximum flow problems, we refer to~\cite{MaxFlow}.
In this sense, we aim at approximating relevant properties of the physical flow between two opposite sides of a~sample by solving flow problems on a~suitable graph network.
\par
\begin{figure}[!ht]
 \centering
   \includegraphics{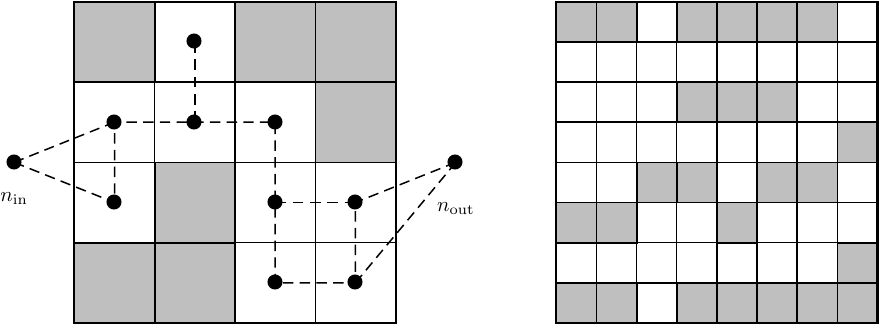} %
    \caption{Left: Derivation of the graph representation from a~pore space illustrated by a~2D $4{\times}4$ pixel image. Nodes are depicted as circles, edges as lines. Gray voxels refer to the solid matrix. The maximum flow value~${f_\mathrm{max}}$ of the depicted graph is one. Right: More complex $8{\times}8$ pixel setup exhibiting ${f_\mathrm{max}}$ of three w.r.t. the horizontal axis and zero w.r.t. the vertical axis.
    }
    \label{FIG:GraphNetwork}
\end{figure}

To do so, we consider the pore space of our segmented image data (stemming from geological specimens) to be a~graph by identifying each voxel as a~node and each neighboring relation of voxels via a~common face as an edge. This procedure is illustrated exemplarily in~\cref{FIG:GraphNetwork} for a~2D structure. For simplicity, all edge weights~$\omega_i$ are assumed to be one. Note that this graph representation still comprises multiple standard characteristics of the pore space. For example, porosity translates to the number of nodes in the graph divided by the total number of voxels; the number of surface elements is approximately determined by $6 \card(N) - 2 \card(E)$. Moreover, we add a~further node $n_\text{in}$ connected to each voxel of the sample's inflow face, as well as $n_\text{out}$ being connected to each voxel of the sample's outflow face, cf.~\cref{FIG:GraphNetwork}. As such, we can approximate the permeability determination problem by a~maximum flow problem through~$G$ between the nodes $n_\text{in},n_\text{out}$. 
\par
For the samples considered in this paper, the computational effort for calculating ${f_\mathrm{max}}$ using Matlab's built-in routine~\texttt{maxflow} with about one second per sample on a~single CPU core, also cf.~\cref{SEC:CNNTrainingAndWorkflowPerformance}, is reasonably small compared to the solution of stationary Stokes equations as described in~\cref{SEC:CompPerformance}. 
By the min-cut max-flow theorem, cf.~\cite{MaxFlow}, the maximum flow problem allows for another interesting interpretation. In the setup introduced within this section, ${f_\mathrm{max}}$ corresponds to the minimal number of edges that need to be deleted from~$G$ to obtain a~disconnected graph such that inflow and outflow faces of the sample are contained in different connected components, cf.~\cref{FIG:GraphNetwork}. Hence, we obtain information about the connectivity of the pore space with respect to the direction of interest. More precisely, ${f_\mathrm{max}}$ classifies structures containing thin channels regarding their restricting effects on fluid flow. This behavior is comprehensible using the geometries illustrated in~\cref{FIG:CNNSamples}: Exhibiting ${f_\mathrm{max}}$ of \num{1230}, the permeability of Subsample~0 is governed by wide channels. On the other hand, exhibiting ${f_\mathrm{max}}$ of~$55$, narrow pore throats hamper the fluid flow of Subsample~9213. 
\par
We further note that for moderate tortuosity, ${f_\mathrm{max}}$ is approximately proportional to the volume of the channel needed to transport the maximum flow through the sample. As such, by exclusively regarding the subgraph of $G$ actively used by the maximum flow, we cut off parts of~$G$ that do not effectively contribute to fluid flow. Hence, the determination of ${f_\mathrm{max}}$ can be understood as approximating the porosity with respect to the pore space participating in fluid transport. That way, we naturally improve the permeability estimation based on \textit{classical} porosities. 
\par
Plotting ${f_\mathrm{max}}$ against the calculated permeabilities $k_\mathrm{cmp}$, we conclude an almost linear relationship between both quantities in the logarithmic scale. Using mean-square linear regression, we obtain the approximate functional relation 
\begin{align}
\label{EQ:MFtoPerm}
k_\mathrm{cmp} \sim 50625 \cdot (f_\mathrm{max})^{1.407} \cdot 10^{-8.183} \quad [\si{D}], 
\end{align}
cf.~\cref{FIG:PhysicsCorrelation}. On the displayed logarithmic scale, this regression yields an approximation quality of $R^2 = 0.8692$, cf.~\cref{SEC:EvaluationMetrics}. Apparently, the derived quantity is strongly correlated to the target permeability $k_\mathrm{cmp}$. Therefore, the permeability estimation using formula~\cref{EQ:MFtoPerm} is considered a~suitable initial guess, which the CNN is able to improve by relating to features extracted from the pore geometry.

\begin{figure}[!ht]
    \centering
    \includegraphics[width=1.0\linewidth]{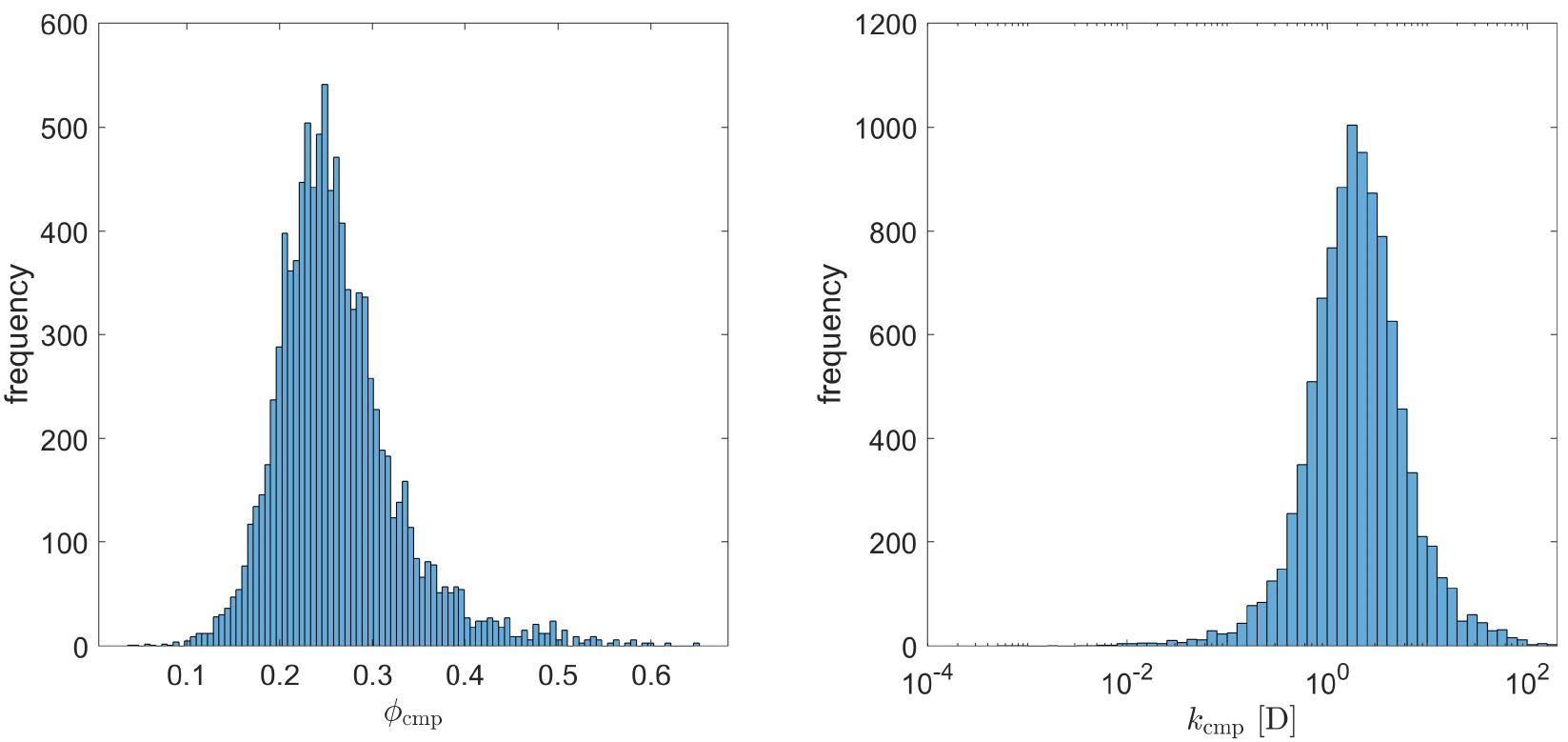}
    \caption{Computed porosity $\phi_\mathrm{cmp}$ and permeability $k_\mathrm{cmp}$ distribution among the total of \num{10000}~Bentheimer samples. Mean porosity is $26.20\%$ with standard deviation $\sigma = 0.067$, mean permeability is \SI{3930}{mD} with $\sigma = \SI{8319}{mD}$.}
    \label{FIG:DataDistribution}
\end{figure}

\section{PhyCNN training and workflow performance}
\label{SEC:CNNTrainingAndWorkflowPerformance}
In this section, information concerning our \mbox{PhyCNN}'s training procedure is provided. Furthermore, we evaluate the quality of our resulting predictions on three different types of sandstone as well as a~challenging artificially deformed series of geometries. Finally, we compare the computational efficiency of our Stokes solver and the \mbox{PhyCNN}. 

\subsection{PhyCNN training}
\label{SEC:CNNTraining}
As the data points obtained by the forward computation are quite sparse for extremely high and low permeability values (\cref{FIG:DataDistribution}), we disregarded all data samples ranging outside the interval of [\SI{50}{mD},~\SI{50}{D}]. As a~result, we improve the quality of predictions within the permeability range where the availability of data is favorable and reliable.
Splitting the remaining data set according to a~$90\% / 10\%$ key, $8\,876$ samples are used for training the network, another 987 for validation. In order to accurately account for the labels covering three orders of magnitude, a~logarithmic transformation was applied before training as done in~\cite{graczyk2020predicting}. Using a~standard mean-squared-error (MSE) loss in the regression as given in~\cref{SEC:EvaluationMetrics}, this results in measuring the error in relative deviation rather than absolute deviation.
\par
The training was performed over 15 epochs using a~stochastic gradient descent (SGD) optimizer with momentum~$0.9$. An almost constant validation loss over the last epochs indicated convergence of the network. Starting from an initial learning rate of~$0.0020$, the step size has been decreased by~$60\%$ every four epochs.  

\subsection{Prediction quality}
\label{SEC:Prediction Quality}
\begin{figure}[!ht]
    \centering
    \includegraphics[width=1.0\linewidth]{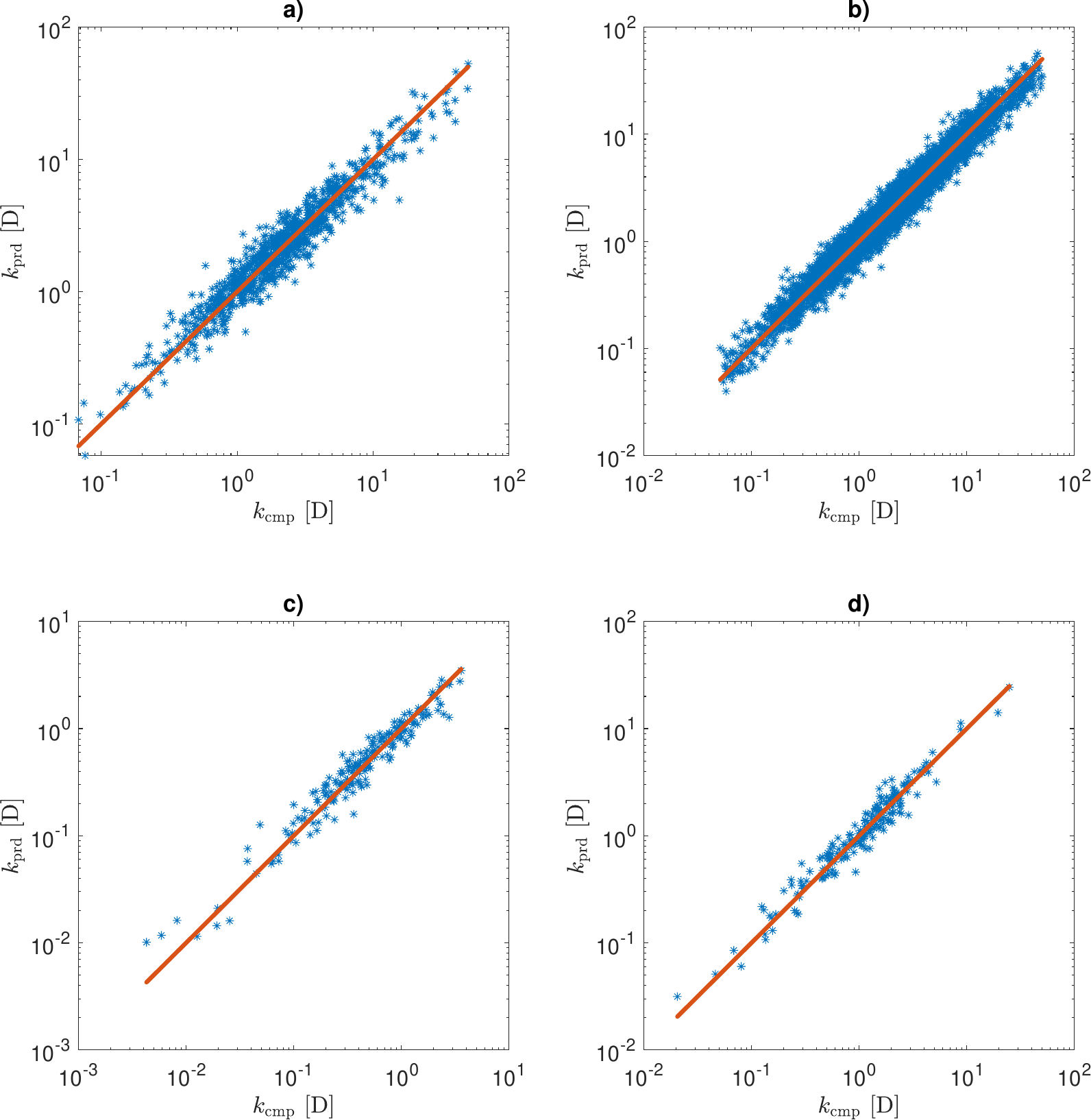}
    \caption{PhyCNN accuracy on validation~a) and training data~b) originating from Bentheimer sandstone. The related $R^2$~values are \num{0.8820}, \num{0.9289} in the natural scale and \num{0.9322}, \num{0.9633} in the logarithmic scaling. Images~c) and~d) show our \mbox{PhyCNN}'s accuracy on the Berea and Castlegate test sets, respectively.}
    \label{FIG:TrainingSuccess}
\end{figure}

The network's predictive performance is illustrated in~\cref{FIG:TrainingSuccess} via regression plots. Our \mbox{PhyCNN} achieved coefficient of determination ($R^2$) values of $96.33\%$ on training data and $93.22\%$ on validation data, cf.~\cref{TAB:PreddictionQuality}. As such, the network accomplished very high accuracy with only weak tendency to overfitting. Furthermore, the plots in~\cref{FIG:TrainingSuccess} show a~low number of outliers, underlining the stability of our approach. 
\par
\begin{table}[ht!]
  \centering
  \begin{tabular}{llcccc}
   \toprule
    & measure & Bentheimer (train.) & Bentheimer (val.)  & Berea &  Castlegate\\
    \midrule
    Data set  &mean $k_\mathrm{cmp}$ [mD] & 3417.7& 3628.8 & 660.9 & 1661.0 \\
    &$\sigma(k_\mathrm{cmp})$ [mD] & 4948.7 & 5233.2 & 657.8 & 2442.2 \\
    \midrule
    PhyCNN($f_\text{max}$)& $R^2$ & 92.89\% & 88.20\% & 89.43\% & 94.55\%      \\
    &$R^2$ log & 96.33\%  & 93.22\% & 94.13\% & 94.51\%    \\
    
    \midrule
    PhyCNN($\phi$-$\sigma$) & $R^2$ & 85.65\% & 73.55\% & 72.81\% & 74.86\%      \\
                       & $R^2$ log & 91.27\%  & 79.57\% & 58.81\% & 75.12\%    \\
    \midrule
     plain CNN & $R^2$ & 78.92\% & 73.11\% & 64.54\% & 62.66\%      \\
                       & $R^2$ log & 84.48\%  & 77.97\% & 58.36\% & 74.12\%    \\
     \midrule
     plain $f_\text{max}$ & $R^2$ & 57.78\%  & 57.67\% & 42.82\% & 56.46\%    \\
     & $R^2$ log & 85.15\%  & 85.03\% & 82.65\% & 81.68\%    \\                  
    \bottomrule
  \end{tabular}
  \caption{Overview prediction quality. In the first block, mean permeability and standard deviation $\sigma(k_\mathrm{cmp})$ are listed for the respective data sets. Subsequently, $R^2$~values in natural and logarithmic scale for the training and validation Bentheimer data as well as for Berea and Castlegate sandstone are presented as achieved by our PhyCNN, i.\,e., the maxflow-informed CNN (PhyCNN($f_\text{max}$)), the porosity/surface-area-informed CNN (PhyCNN($\phi$-$\sigma$)), and a~plain CNN without additional inputs. Finally, the additional input quantity~$f_\text{max}$ is considered in a~power-law (linear fitting applied to logarithmic variables) regression model (plain~$f_\text{max}$).}
  \label{TAB:PreddictionQuality}  
\end{table}

In order to further stress the robustness, we additionally validate our \mbox{PhyCNN} on samples originating from different types of sandstone that were not used for training. Using the data provided by~\cite{BentheimerDataSet,BentheimerPublication}, 200~additional subsamples were extracted from each Berea and Castlegate \muCT~scan, see~\cref{TAB:SampleCharacteristics}, using the same methods as described in~\cref{SEC:Datapreparation}. Achieving $R^2$ values of 89.43\% and 94.13\% (logarithmic) on Berea as well as 94.55\% and 94.51\% (logarithmic) on Castlegate, the networks proves excellent generalization properties across different types of pore geometries, cf.~\cref{FIG:TrainingSuccess}, \cref{fig:PoreSpace}, and~\cref{TAB:SampleCharacteristics}. As such, the network seems in fact to perform slightly better on Berea and Castlegate data samples than on the validation data set of Bentheimer. However, \cref{TAB:PreddictionQuality} marks a~very high standard deviation for the permeability in Bentheimer compared to the other sandstone types. As such, the latter data sets comprise less heterogeneity facilitating the network's prediction. We further emphasize that both Berea and Castlegate validation sets have not been screened to match the trained range [\SI{50}{mD},~\SI{50}{D}]. More precisely, eleven Berea data samples exhibit a~permeability $k_\mathrm{cmp}$ below \SI{50}{mD} as well as two Castlegate data samples.
\par
Moreover, \cref{TAB:PreddictionQuality} lists the prediction quality indices of other related CNNs for comparison. To clearly and consistently separate the different approaches investigated, we refer to the PhyCNN described in~\cref{SEC:CNN} more precisely as PhyCNN($f_\text{max}$) within this comparison. In case of the PhyCNN($\phi$-$\sigma$), the general structure depicted in~\cref{FIG:PhyCNNStructure} is maintained while replacing the maxflow input variable by the porosity~$\phi$ and surface area~$\sigma$, and retraining the network, cf.~\cref{FIG:PhysicsCorrelation}. As the data of~\cref{TAB:PreddictionQuality} show, the PhyCNN($\phi$-$\sigma$) cannot reach the accuracy of our PhyCNN($f_\text{max}$). Moreover, robustness appears significantly lower since accuracy on other sandstone types than the training data decreases rapidly. Especially for Berea sandstone, the results are poor, which might be a~direct consequence of the significantly different pore-space characteristics in comparison to Bentheimer, cf.~\cref{TAB:SampleCharacteristics}. Furthermore, we also compare to the analogous network without any additional input quantities (plain~CNN), resulting in slightly worse predictions than the PhyCNN($\phi$-$\sigma$). Finally, we note that both of these CNNs provide less accurate predictions on validation samples than $f_\text{max}$ alone (plain~$f_\text{max}$) being fitted to the Bentheimer training data set using a~power law approach and measured with respect to $R^2$-log. As such, we conclude that $f_\text{max}$ already contains highly relevant information regarding the samples' permeability delivering more accurate predictions in the logarithmic measure than standard CNN approaches. However, our PhyCNN($f_\text{max}$) improves the results of pure $f_\text{max}$ by extracting further information directly from the pore-space, cf.~\cref{FIG:PhyCNNStructure}.
\par

\begin{figure}
    \centering
    \includegraphics[width=0.9\textwidth]{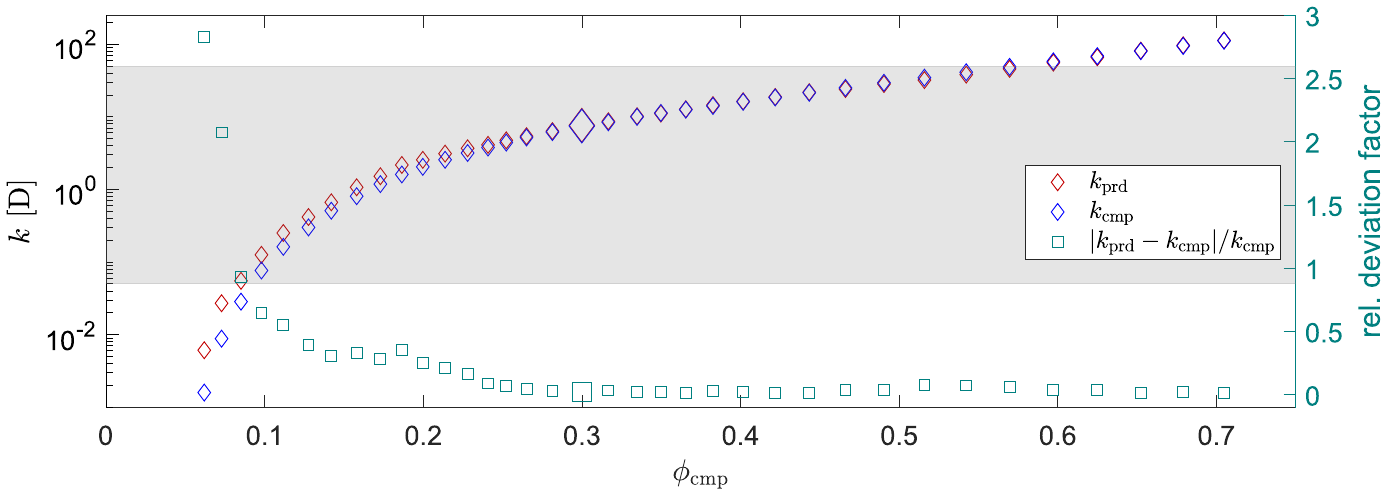}
    \caption{Testing generalization ability on artificially distorted data samples. \mbox{PhyCNN} prediction performance on eroded and dilated pore spaces exhibiting a~large range of porosities $\phi_\mathrm{cmp}$. For each manipulated geometry, Stokes simulation data~$k_\mathrm{cmp}$ (red) and CNN prediction $k_\mathrm{prd}$ (blue) are compared. Furthermore, we provide relative prediction errors. Markers referring to the original, not manipulated data sample are increased in size. The permeability range spanned by our \mbox{PhyCNN}'s training data set is highlighted in gray. }
    \label{FIG:Dilute}
\end{figure}

Finally, we investigate the generalization limits of our \mbox{PhyCNN} by subjecting it to an~artificially distorted data set covering a~challengingly wide permeability and porosity range. To achieve that, we applied a~level-set-based algorithm already used in~\cite{Gaerttner2020b} to erode or dilate the pore space of Subsample~0 displayed in~\cref{FIG:CNNSamples}. More precisely, the pore space is contracted with a~uniform level-set velocity directed perpendicularly to the pore walls until all flow channels collapse, i.\,e.\ the data sample becomes impermeable. The same number of deformation steps is also used to expand the pore volume of Subsample~0. Each of those steps corresponds to a~constriction/expansion of the pore space by a~single additional voxel layer on average.
\par
Subsequently, we compute the permeabilities~$k_\mathrm{cmp}$ using the Stokes solver and $k_\mathrm{prd}$ using the \mbox{PhyCNN} of the series of pore spaces and compare the results in~\cref{FIG:Dilute}. The data show an almost perfect match for expanded pore spaces even for porosities up to 70\%. Using linear interpolation, we estimate $k_\mathrm{cmp}$ to exceed the trained permeability range~[\SI{50}{mD},~\SI{50}{D}] beyond $\phi_\mathrm{cmp}=57.33\%$. Therefore, we conclude that our \mbox{PhyCNN} is capable of properly characterizing also highly permeable samples. On the other hand, predictions remain reasonably accurate for strongly confined geometric structures. Three samples in~\cref{FIG:Dilute} show porosities below $9.08\%$, which approximately refers to the lower end of the trained permeability range for this example. These exhibit ${f_\mathrm{max}}$ of~4, 16, and 30, respectively, referring to an almost disconnected pore space. Since the flow channels narrow in these cases to only very few voxels in diameter, errors from the \muCT~scan as well the discretization of Stokes equations~\eqref{EQ:StokesProblem} may become non-negligible. As such, these samples may leave the current operating limits of digital rock physics, resulting in a~significant systematic overestimation of permeability compared to lab experiments~\cite{SaxenaDRPlimit}.

\subsection{Computational performance}
\label{SEC:CompPerformance}
In this section, we provide computational performance indicators for the forward simulations as performed using the method described in~\cref{SEC:ForwardSim} as well as the ones obtained for estimations using the \mbox{PhyCNN}. 
Subsequently, we compare the forward simulation run times for the generation of \num{10000}~data samples including both permeability computation approaches on the same compute cluster to estimate the actual speed up. All subsequent specifications of computation times refer to the wall time.
\par
Each of the \num{10000}~forward simulations on Bentheimer sandstone was performed with classical Taylor--Hood elements on voxels ($\ell{=}1$, cf.~\cref{SEC:DiscChoice}) in parallel on 50~compute nodes of the Emmy compute cluster at RRZE~\cite{EmmyCluster}, each being equipped with two Intel\textregistered Xeon~2660v2 \enquote*{Ivy Bridge} processors (10~physical cores per chip, i.\,e., $50{\cdot}2{\cdot}10{=}1000$ cores total, no hyper-threading) and \SI{64}{GB} of RAM.
The linear solver (MINRES) converged within 969~iterations on average with a~relative tolerance of~\num{1.0E-6} in the norm~\mbox{$\|\cdot\|_{\mathcal{P}^{-1}}$}. Furthermore, meshing, assembly, and solution were performed within $2.35$, $0.44$, and $33.95$~seconds on average per solver call, respectively. As such, the labeling procedure of all \num{10000}~data samples was completed within roughly \num{100}~hours. We conclude that the forward simulation effort of our Stokes approach is roughly comparable to the LBM-based simulations applied in related publications, cf.~\cite{RapidEstimate}.
\par
Operating on a~graphics cluster featuring two Intel\textregistered Xeon~E5-2620v3 (6~cores each), two Nvidia{\textregistered} Geforce Titan~X~GPUs, and 64~GB of~RAM, the \mbox{PhyCNN}'s training process terminated within two hours. Using 10~CPU cores and a~single graphics chip on this machine, the permeabilities of the~\num{10000} elements data set of Bentheimer sandstone is estimated within \num{1294}~seconds by our \mbox{PhyCNN}. More precisely, \num{867}~seconds on CPU were spent on the graph flow problems related to~${f_\mathrm{max}}$ as described in~\cref{SEC:MaxFlow} while \num{367}~seconds were required to perform the subsequent network inferencing on GPU. Estimating the time required to solve the related Stokes problem~\eqref{EQ:StokesProblem} extrapolated from a~subset of ten data samples indicates a~total time of approximately \num{1469}~hours. Accordingly, we conclude an acceleration factor of~\num{4087} by using our \mbox{PhyCNN} on the stated hardware configuration and data set. 
\par
From the data presented above, we infer that the time invested in the network's training procedure including the calculation of~${f_\mathrm{max}}$ for the complete database is compensated after about \num{15}~data samples computed beyond the training and validation data sets. This strongly underlines the capability of our approach to pose a~time-saving yet accurate alternative to flow simulation-based permeability estimators on large data sets.

\section{Summary and conclusions}
\label{SEC:SummaryandConclusions}

In this work, we demonstrated the feasibility of direct numerical simulation (DNS) for flow through porous media to generate a~library of computed permeability labels from 3D~images acquired using specimens of natural rocks. Our distributed-parallel stationary Stokes solver achieved computational efficiency comparable to classical lattice Boltzmann method (LBM) implementations while successfully addressing convergence challenges that typically appear on complex 3D~geometries including poorly resolved subdomains.
\par
As a~result of computations with the proposed numerically robust forward simulation algorithm, an unbiased (by computational artifacts) data set was constructed to train a~convolutional neural network for accelerated permeability predictions. An~easily-computable graph-based characteristic quantity of the pore space, namely the maximum flow value, was introduced to the neural network as another novel development in our work. As a~result, this methodology enabled our machine learning model to achieve an~$R^2$~value of $93.22\%$ on the validation data set. Moreover, similar prediction qualities were found for types of sandstone rock that are different from the training samples, as well as for artificially generated voxel sets. These observations underline the robustness of our artificial intelligence augmented permeability estimation approach. 
\par
In a~one-on-one comparison before accounting for the training-investment-related computational costs, the neural-network-based permeability estimation approach delivered a~speed-up in excess of \num{4000} fold. On the other hand, computational results indicate that the proposed approach exceeds the performance of a~purely DNS-based workflow beyond approximately 15~data samples when the training-related computational costs are accounted for. This indicates that the proposed artificial intelligence augmented permeability estimation workflow is viable for real-life digital rock physics applications.       
\par
Finally, we note that the refined methodology presented in this paper may generalize to larger sample sizes approaching the REV-scale. Recent publications suggest hierarchical multiscale neural networks~\cite{Santos2021MultiscaleNetwork} to alleviate the memory requirements of the training and inference procedure. Merging both approaches holds the potential of leveraging our current results to larger scale geological samples.

\section*{Data availability}

The CNN code used in this paper is available as part of the porous media numerical toolkit RTSPHEM~\cite{RTSPHEM} on Github.

\section*{Notation}

\subsection*{Symbols}
\begin{tabularx}{\linewidth}{@{}lL@{}}
$A_\mathrm{cmp}$ & Computed interior surface area of \muCT~scan [$\si{\milli\meter}^2$].\\
$A_\mathrm{spec}$ & Specific (w.r.t. material volume) interior surface area [$\si{\milli\meter}^{-1}$].\\
$\alpha$ & Parameter in \texttt{LeakyReLU} nonlinearity. \\
${f_\mathrm{max}}$ & Maximum flow value. \\
$k_\mathrm{ana}$ & Analytical permeability (all permeabilities in darcies~D, millidarcies~mD).\\
$k_\mathrm{cmp}$ & Computed permeability, cf.~\eqref{EQ:PermEstFormula}.  \\
$k_\mathrm{exp}$ & Experimental permeability.\\
$k_\mathrm{prd}$ & Predicted permeability.\\
$\ell$ & Polynomial degree along principle axes ($\IQ_\ell$ is the local space of polynomials of degree at most~$\ell$ in each variable).\\
$\Omega$ & Pore space, domain~$\Omega\subset\IR^3$ of the Stokes equations~\eqref{EQ:StokesProblem}.\\
$p$      & Hydrostatic pressure on the pore scale (dimensionless), solution $p:\Omega\rightarrow\IR$ of the Stokes equations~\eqref{EQ:StokesProblem}.\\
$\phi_\mathrm{cmp}$  & Computed porosity. \\
$\phi_\mathrm{exp}$  & Experimental porosity. \\
$R^2$    & R-squared value, coefficient of determination.\\
$\Re$    & Reynolds number (definition is irrelevant here, since $\Re$ cancels out in the computation for~$k_\mathrm{cmp}$, cf.~\eqref{EQ:PermEstFormula}).\\
$\sigma$    & Standard deviation.\\
$\vec{u}$ & Solenoidal fluid velocity on the pore scale (dimensionless), solution $\vec{u}:\Omega\rightarrow\IR^3$ of the Stokes equations~\eqref{EQ:StokesProblem}.\\
$n$ & Node $n\in N$ in an~undirected graph~$G(N,E,\omega)$, cf.~\cref{SEC:MaxFlow}.\\
\end{tabularx}
\subsection*{Abbreviations}
\begin{tabularx}{\linewidth}{@{}lL@{}}
CNN    & Convolutional neural network.\\
DNS    & Direct numerical simulation. \\
DOF    & Degrees of freedom.\\
FFN    & Feed forward network.\\
LBM    & Lattice Boltzmann method.\\
\muCT  & Microcomputed tomography.\\
MSE    & Mean squared error. \\
PhyCNN & Physics-informed convolutional neural network.\\
ReLU   & Rectified linear unit.\\
REV    & Representative elementary volume.\\
SGD    & Stochastic gradient descent.
\end{tabularx}

\section*{Acknowledgments}
S.~G\"arttner and A.~Meier were supported by the DFG Research Training Group 2339 Interfaces, Complex Structures, and Singular Limits.\\
N.~Ray was supported by the DFG Research Training Group~2339 Interfaces, Complex Structures, and Singular Limits and the DFG Research Unit~2179 MadSoil.\\
F.~Frank was supported by the Competence Network for Scientific High Performance Computing in Bavaria (KONWIHR).
\\[\baselineskip]
We further thank Martin Burger for insightful discussions and Fabian Woller for assisting with the MFEM~implementation.

\clearpage
\bgroup
\emergencystretch 3em 
\printbibliography[heading=bibintoc]
\egroup

\end{document}